\documentclass{article}
\usepackage{iclr2027,times}

\usepackage{amsmath,amsfonts,bm}
\usepackage{caption}
\usepackage{booktabs}
\usepackage{multirow}
\usepackage{siunitx} 
\usepackage{amsmath}
\usepackage{subcaption}
\usepackage{setspace} 

\def\eqref#1{equation~\ref{#1}}

\def\1{\bm{1}}

\DeclareMathAlphabet{\mathsfit}{\encodingdefault}{\sfdefault}{m}{sl}
\SetMathAlphabet{\mathsfit}{bold}{\encodingdefault}{\sfdefault}{bx}{n}

\usepackage{amsmath}
\usepackage{amssymb}
\usepackage{array}
\usepackage{booktabs}
\usepackage{graphicx}
\usepackage{wrapfig}
\usepackage{multirow}
\usepackage{makecell}
\usepackage{enumitem}
\usepackage{tikz}
\usepackage{colortbl}
\usetikzlibrary{arrows.meta,positioning}
\usepackage{hyperref}
\hypersetup{hidelinks}
\usepackage{url}

\setlist[itemize]{leftmargin=*, itemsep=2pt, topsep=3pt}
\newcommand{\sgdetpp}{FFBL-Coop}
\newcommand{\method}{\sgdetpp}

\definecolor{best}{RGB}{255,180,180}
\definecolor{second}{RGB}{255,230,180}
\definecolor{third}{RGB}{255,255,200}
\definecolor{besterror}{RGB}{210,230,255}
\definecolor{seconderror}{RGB}{228,240,255}
\definecolor{thirderror}{RGB}{240,247,255}

\newcolumntype{C}[1]{>{\centering\arraybackslash}p{#1}}
\newcolumntype{L}[1]{>{\raggedright\arraybackslash}p{#1}}

\title{\sgdetpp{}: Association-Decoupled Cooperative 3D Multi-Object Tracking}
\author{
	Haoxin Wu\textsuperscript{1}\textsuperscript{*},
	Xiaokai Bai\textsuperscript{1}\textsuperscript{*},
	\\[2mm]
	\textsuperscript{1}College of Information Science and Electronic Engineering, Zhejiang University.
	\\
	\textsuperscript{*} Equal contribution, \textsuperscript{$\dagger$} Corresponding author.
	\texttt{email:shawnnnkb@gmail.com}
}
\iclrfinalcopy
\begin{document}
	
	\maketitle
	\fancyhead[L]{Under review as a conference paper at ICLR 2027}
	
\vspace*{-15pt}
\begin{abstract}
	Cooperative 3D tracking must integrate complementary observations
	across agents and time while maintaining consistent identities.
	When evidence integration and identity inheritance share
	a matching decision, errors arising from cross-view appearance
	differences and spatial misalignment can compromise both
	feature fusion and track continuity.
	We propose \sgdetpp{}, a \emph{fuse first, bind later} framework
	that separates instance admission from identity management.
	Confidence-ranked Slot Admission (CSA) allocates cooperative queries
	to available ego slots using confidence and spatial proximity.
	Unified Representation Aggregation (URA) uses cooperative
	semantic features and aligned anchors to guide ego-feature
	retrieval, refining the augmented query bank within a shared
	transformer decoder.
	After refinement, Cooperative-Priority Identity Anchoring
	(CPIA) combines learned association with persistent mappings
	to establish accepted identity assignments across frames.
	A shared codebook reduces transmitted payload while retaining
	AP and AMOTA close to the uncompressed variant.
	\sgdetpp{} achieves AMOTA/AP of $0.611/0.548$ on V2X-Seq
	and $0.688/0.653$ on Griffin-25M.
	Code will be released.
\end{abstract}

\vspace{-10pt}
\section{Introduction}
\label{sec:introduction}

Cooperative 3D tracking extends ego perception through observations from nearby vehicles, roadside infrastructure, or aerial platforms. These complementary views reveal objects obscured or outside the ego field of view. Exploiting this coverage over time requires integrating incoming evidence while preserving identities across changes in viewpoint and visibility.

Intermediate fusion supports evidence integration by exchanging learned representations, such as bird's-eye-view (BEV) features~\citep{xu2022v2xvit,xu2022cobevt}. To reduce communication cost, recent systems exchange sparse object queries and combine cross-agent association with feature aggregation~\citep{yu2025univ2x,zhong2025cooptrack,wang2025sparsecoop}. Within this sparse formulation, SparseCoop and Long-SCOPE accommodate unmatched queries, and SparseCoop further refines representations using ego image features~\citep{wang2025sparsecoop,wang2026longscope}. Incorporating these representations into a recurrent tracker also requires deciding which trajectories the updated hypotheses continue.

\begin{figure}[t]
	\centering
	\includegraphics[width=\textwidth]{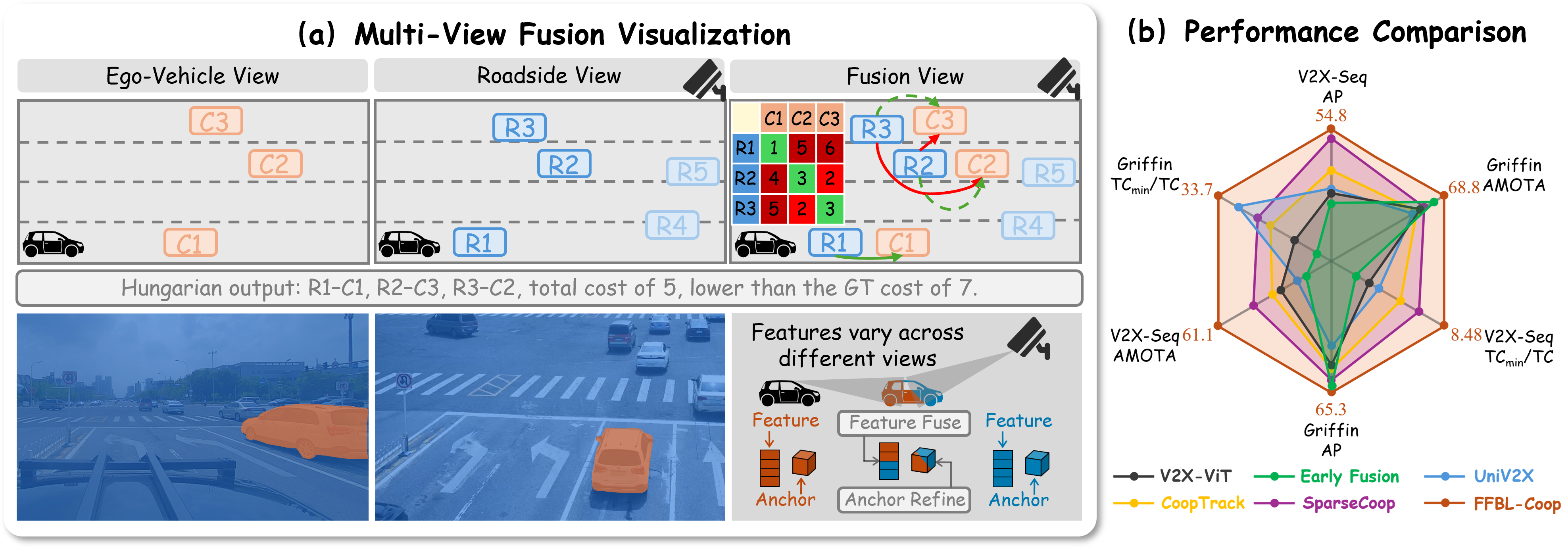}
\caption{
	\textbf{(a) Cross-view matching challenges.}
	\textit{Top:} Geometric discrepancies can make the
	minimum-cost assignment inconsistent with ground-truth
	object correspondences.
	\textit{Bottom:} The same object has different feature
	representations and anchor estimates across agent views.
	\textbf{(b) System comparison.}
	Detection, tracking, and communication performance on
	V2X-Seq and Griffin-25M.
	Annotations show AP/AMOTA in percent and transmission
	rates in kB/s.
}
	\label{fig:motivation}
	\vspace{-12pt}
\end{figure}

Cooperative integration and temporal tracking impose different
requirements on association.
Specifically, integration must admit useful cooperative evidence
despite cross-view appearance differences and alignment errors,
whereas tracking must preserve identity across frames.
Yet when one cross-agent match controls both query placement
and identity inheritance, a correspondence error can compromise
both representation updates and trajectory continuity
(Figure~\ref{fig:motivation}(a)).
Crucially, uncertainty about which existing track an observation
corresponds to does not necessarily eliminate its value
for detection.
This distinction motivates our principle:
\emph{integrate and refine cooperative evidence first,
	then bind the resulting hypotheses to persistent trajectories}.

Following this principle, \sgdetpp{} separates evidence admission
from ego-track identity assignment through a
\emph{fuse first, bind later} pipeline.
CSA selects cooperative queries by confidence and spatial proximity
and places them in available ego slots without assigning ego IDs.
URA then uses their semantic features and aligned anchors
to guide ego-feature retrieval and refine the augmented bank.
Only after final decoding does CPIA resolve identities
using refined ego states and persistent cross-agent mappings,
independently of admission weights.
Learned association links queries to existing tracks,
accepted assignments and new-track identities are recorded
in persistent ID mappings.

Figure~\ref{fig:framework} summarizes the complete pipeline,
from cross-agent alignment and cooperative admission to
shared decoding and identity binding.
It shows how cooperative queries enter the recurrent ego bank
and are refined alongside native ego queries.
Valid out-of-region queries follow a separate box-output path
to retain coverage beyond local integration, while a shared
codebook compresses the cooperative features used for refinement.
Experiments on V2X-Seq and Griffin-25M evaluate detection,
tracking, and communication performance
(Figure~\ref{fig:motivation}(b)).
Controlled ablations examine whether separating identity
management from feature admission improves tracking and
how persistent ID mappings contribute to continuity.

Our main contributions are:
\begin{itemize}
	\item We formulate cooperative query tracking as a
	\emph{fuse first, bind later} process: cooperative evidence
	is admitted and refined before identity assignment.
	This separates the decision to integrate an observation
	from the decision of which trajectory it continues.
	
	\item We instantiate this principle through three stages.
	CSA ranks queries by confidence and proximity for admission,
	URA uses cooperative semantics to guide ego-feature retrieval,
	and CPIA combines association with persistent mappings
	to maintain identities across slot changes.
	
	\item Experiments on V2X-Seq and Griffin-25M demonstrate the effectiveness of \textsc{\sgdetpp{}} for cooperative 3D detection and tracking.
	Codebook compression reduces payload while retaining
	AP and AMOTA close to uncompressed transmission.
\end{itemize}

\vspace*{-5pt}

\section{Related Work}
\label{sec:related}

\vspace*{-5pt}

\subsection{3D Multi-Object Tracking}
\label{sec:MOT}

Tracking-by-detection methods maintain trajectories by
associating per-frame detections using geometric and motion
cues~\citep{weng2020ab3dmot,yin2021centerpoint}.
A complementary line of work learns object representations
through queries.
Specifically, query-based 3D detectors such as
DETR3D~\citep{wang2022detr3d} and PETR~\citep{liu2022petr}
aggregate multi-view image features into sparse object queries.
Extending this representation across time,
StreamPETR~\citep{wang2023streampetr} and
Sparse4D~\citep{lin2024sparse4d} incorporate historical
information into object-centric representations,
providing a foundation for recurrent updates.

Beyond temporal feature aggregation,
query propagation also supports identity maintenance.
MOTR~\citep{zeng2022motr} propagates track queries across
video frames, while DQTrack~\citep{park2023dqdetr} separates
object and track queries to address representation conflicts
between detection and tracking in 3D.
ADA-Track~\citep{ding2024adatrack} alternates detection
and association within decoder layers.
In a cooperative setting, recurrent updates additionally
incorporate queries maintained by another agent.
\sgdetpp{} separates cooperative-query admission from ego-track
identity inheritance using persistent cross-agent ID mappings
to support continuity across slot changes.

\subsection{Cooperative Perception}
\label{sec:COOPERATIVE}

	\vspace{-2pt}

Cooperative perception extends an agent's coverage by
incorporating observations from other viewpoints.
Within this paradigm, intermediate fusion exchanges learned
features to integrate complementary evidence.
F-Cooper~\citep{chen2019fcooper} studies feature-level fusion,
while V2X-ViT~\citep{xu2022v2xvit} and
CoBEVT~\citep{xu2022cobevt} use attention to aggregate
multi-agent representations.
To make communication more selective,
When2com~\citep{When2com} and Where2comm~\citep{hu2022where2comm}
use learned communication decisions and spatial confidence
maps, respectively.
Complementing these efficiency-oriented designs,
CoAlign~\citep{lu2023coalign} addresses pose uncertainty
through agent and object alignment.
At the object level, UniV2X~\citep{yu2025univ2x} exchanges
sparse queries for cooperative perception and planning.

Building on sparse representations, several methods combine
cross-agent association, ture refinement, and temporal
propagation~\citep{zhong2025cooptrack,wang2025sparsecoop,wang2026longscope}.
CoopTrack uses graph-based affinities and Hungarian matching
to aggregate paired queries.
SparseCoop combines geometric and appearance cues,
refining queries with ego-image context.
Long-SCOPE uses contextual descriptors and optimal-transport
matching for long-range association.
Despite advances in matching and refinement, coupling identity
inheritance to query allocation can let matching errors affect
both evidence integration and track continuity.
\sgdetpp{} addresses this coupling through a
\emph{fuse first, bind later} pipeline that separates feature admission from identity inheritance and records accepted assignments as persistent cross-agent ID mappings.

	\begin{figure*}[t]
		\centering
		\includegraphics[width=\textwidth]{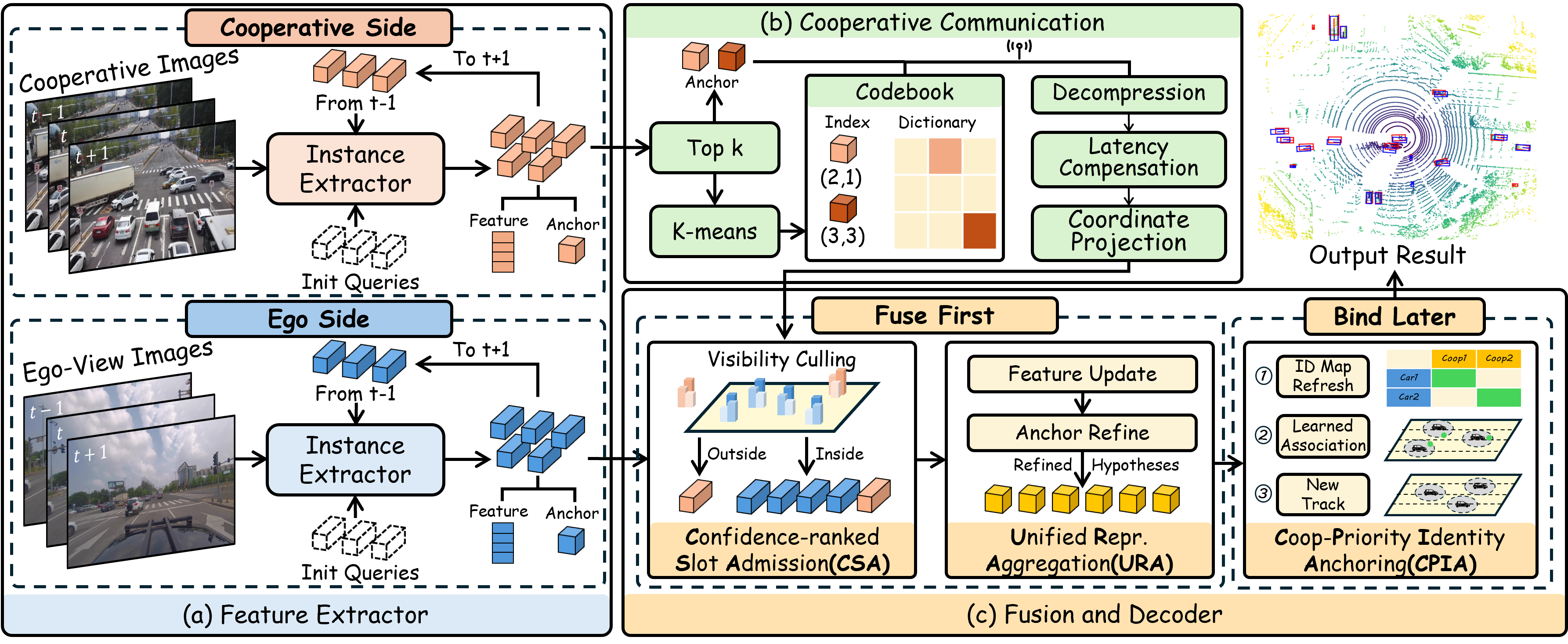}
		\caption{
			\textbf{\method{} framework.}
			Cooperative queries aligned to the ego frame through motion
			compensation and coordinate transformation.
			Following \emph{fuse first, bind later}, CSA admits cooperative
			evidence into available slots ,
			and URA uses cooperative semantics to guide ego-feature
			retrieval and refinement.
			CPIA assigns identities through learned association and ID mappings.
		}
		\label{fig:framework}
		\vspace{-10pt}
	\end{figure*}

	\section{Method}
	\label{sec:method}
	
	\vspace{-4pt}
	
\subsection{Overview}
\label{sec:overview}

	\vspace{-4pt}

\method{} follows a \emph{fuse first, bind later} pipeline: cooperative evidence is integrated into the ego query bank before the hypotheses receive track identities (Figure~\ref{fig:framework}). Each query is represented by $(f,a,s,\mathrm{id})$, where $f\in\mathbb{R}^{C}$ is an instance feature, $a$ is a 3D anchor encoding the object state, $s$ is a confidence score, and $\mathrm{id}$ is the track identifier maintained by the originating agent.

At each communication step, the cooperative agent provides a query set
\begin{equation}
	Q_t^{\mathrm{coop}}
	=
	\bigl\{
	(f_j^{\mathrm{coop}},a_j^{\mathrm{coop}},
	s_j^{\mathrm{coop}},\mathrm{id}_j^{\mathrm{coop}})
	\bigr\}_{j=1}^{N_{\mathrm{coop}}}.
	\label{eq:qcoop}
\end{equation}
The feature component is transmitted through a shared codebook. The ego agent reconstructs the features and aligns the cooperative states to its current coordinate frame and timestamp (Sec.~\ref{sec:comm}).

Geometrically valid aligned queries whose centers project
within an ego camera image form the cooperative candidate pool
(Sec.~\ref{sec:supp_integration_region}).
CSA allocates selected candidates to available slots,
URA refines their features with ego observations,
and CPIA resolves identities.
Valid queries outside all ego camera images bypass decoding
and contribute boxes through a separate output path.
The final output combines bank predictions and these cooperative detections:
\begin{equation}
	Y_t
	=
	\bigl\{
	(\hat{a}_i,\hat{s}_i,\widehat{\mathrm{id}}_i)
	\bigr\}_{i=1}^{N_t^{\mathrm{out}}},
	\label{eq:yout}
\end{equation}
where $N_t^{\mathrm{out}}$ is the number of output hypotheses, distinct from the fixed bank capacity $N$. The integration and identity rules are described in Sec.~\ref{sec:decoupled-integration}.

\subsection{Agent-Specific Perception Stacks}
\label{sec:single}

\begin{wrapfigure}{l}{0.5\columnwidth}
	\centering
	\vspace{-10pt}
	\includegraphics[width=\linewidth]{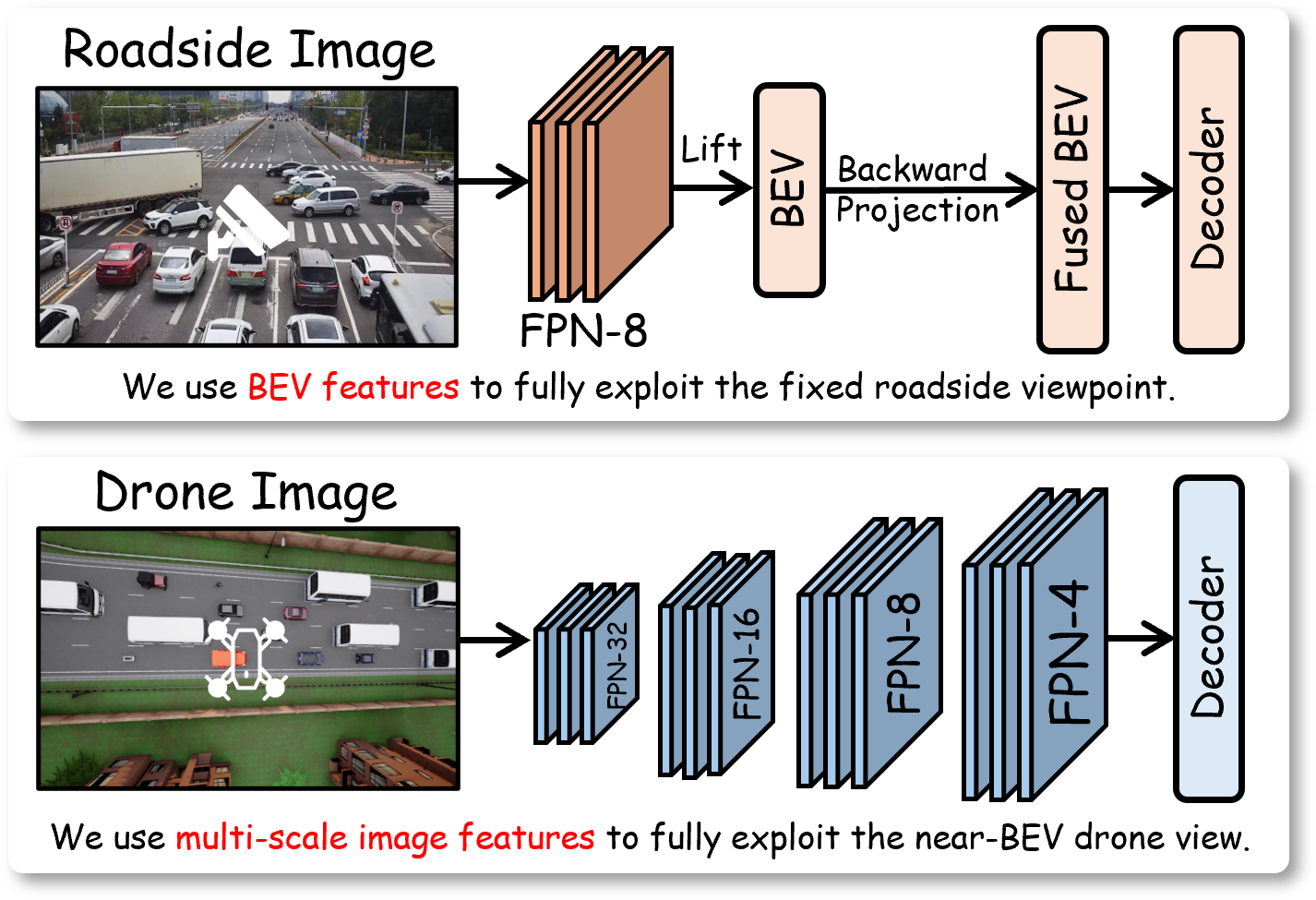}
	\caption{
		\textbf{Agent-specific perception stacks.}
		(a) Roadside queries aggregate evidence from constructed
		BEV features.
		(b) Aerial queries retrieve evidence from multi-scale
		FPN features.
	}
	\label{fig:frontends}
	\vspace{-20pt}
\end{wrapfigure}

The perception frontends use specific spatial representations and expose the same query interface $(f,a,s,\mathrm{id})$ for cooperation. Roadside units construct BEV features, while ego vehicles and aerial agents sample image features through projected 3D anchors. Coordinate alignment and feature adaptation are applied before cross-agent integration.

\vspace*{-5pt}

\paragraph{Feature Encoding.}
Each agent processes its images with a backbone and feature pyramid network (FPN), producing multi-scale features $F_t^{I}$. For ego vehicles and aerial agents, queries project their 3D anchors into the camera views and aggregate features around the projected locations. This readout uses camera geometry without requiring the image plane to approximate a BEV layout.

For roadside units, LSS~\citep{philion2020lift} and deformable refinement construct BEV features $F_t^{\mathrm{BP}}$ from multi-scale image features. An image-conditioned hypernetwork predicts spatially varying modulation parameters $(\gamma,\beta)$ and a gate $g$, yielding the final representation:
\begin{equation}
	F_t^{\mathrm{BEV}}
	=
	(1-g)\odot F_t^{\mathrm{BP}}
	+
	g\odot
	\bigl(
	\gamma\odot F_t^{\mathrm{BP}}+\beta
	\bigr),
	\label{eq:igdr}
\end{equation}
where $\odot$ denotes element-wise multiplication, with broadcasting over feature dimensions where required. The gate controls the contribution of the image-conditioned modulation.

\paragraph{Sparse Queries and Recurrent Decoding.}
Each agent maintains a recurrent bank $B_{t-1}$ of $N$ query slots. A decoder layer updates the current queries, followed by cross-attention to the previous bank, producing $B_t^{\mathrm{temp}}$. Subsequent decoder layers update query features and anchors:
\begin{equation}
	\bigl(f_i^{\ell+1},a_i^{\ell+1}\bigr)
	=
	\mathcal{D}_{\ell}
	\bigl(
	f_i^{\ell},a_i^{\ell},F_t,B_t^{\mathrm{temp}}
	\bigr),
	\label{eq:single-decoder}
\end{equation}
where $\ell$ indexes decoder layers and $F_t$ denotes the spatial features supplied by the frontend. This recurrent object-centric processing follows the temporal-query design of existing sparse perception models~\citep{lin2024sparse4d,wang2023streampetr}. In the ego pipeline, admission and identity updates occur after the temporal update. Subsequent layers jointly refine cooperative and native ego queries.

\subsection{Communication and Cross-Agent Alignment}
\label{sec:comm}

\paragraph{Codebook Compression.}
CodeFilling~\citep{hu2024codefilling} explores codebook-based
communication for collaborative perception.
Here, we apply vector quantization~\citep{linde1980vq}
to sparse cooperative instance features.
We construct a shared codebook
$\mathcal{C}\in\mathbb{R}^{K_{\mathrm{code}}\times C}$
by applying $k$-means to training-set features,
where $K_{\mathrm{code}}$ denotes the codebook size.
Each cooperative feature is encoded by its nearest codeword:
\begin{equation}
	z_j
	=
	\arg\min_{k\in\{1,\ldots,K_{\mathrm{code}}\}}
	\left\|
	f_j^{\mathrm{coop}}-\mathcal{C}_{k}
	\right\|_2^2.
	\label{eq:vq-enc}
\end{equation}
The transmitted query contains
$(z_j,a_j^{\mathrm{coop}},s_j^{\mathrm{coop}},
\mathrm{id}_j^{\mathrm{coop}})$.
The ego agent reconstructs the feature by table lookup,
$\hat{f}_j^{\mathrm{coop}}=\mathcal{C}_{z_j}$,
while retaining the object state, confidence, and source track ID.

For queries integrated into the ego bank, reconstructed features provide object-level semantic cues for initialization and aggregation. Subsequent URA refinement incorporates local image evidence from ego spatial features. This use of cooperative features motivates codebook compression.

\begingroup
\setlength{\abovedisplayskip}{4pt plus 1pt minus 1pt}
\setlength{\belowdisplayskip}{4pt plus 1pt minus 1pt}
\setlength{\abovedisplayshortskip}{2pt plus 1pt}
\setlength{\belowdisplayshortskip}{3pt plus 1pt minus 1pt}

\paragraph{Latency and Coordinate Alignment.}
Before integration, the ego agent compensates for timestamp differences and transforms cooperative object states into its coordinate frame. Let $\Delta t=t^{\mathrm{ego}}-t^{\mathrm{coop}}$. Under a constant-velocity approximation, the cooperative center is updated as
\begin{equation}
	p_j^{+}
	=
	p_j^{\mathrm{coop}}
	+
	v_j^{\mathrm{coop}}\Delta t,
	\label{eq:motion-compensation}
\end{equation}
where $p_j^{\mathrm{coop}}$ and $v_j^{\mathrm{coop}}$ are expressed in the cooperative frame. Applying
$T_{\mathrm{coop}\rightarrow\mathrm{ego}}$
to the compensated state yields the aligned anchor
$\tilde{a}_j^{\mathrm{coop}}$, with its center and orientation transformed and velocity rotated into the ego frame.

A learned linear adapter conditions reconstructed features on the relative agent rotation:
\begin{equation}
	\tilde{f}_j^{\mathrm{coop}}
	=
	A_{\mathrm{feat}}
	\left(
	\left[
	\hat{f}_j^{\mathrm{coop}};
	\mathrm{vec}
	\bigl(R_{\mathrm{coop}\rightarrow\mathrm{ego}}\bigr)
	\right]
	\right),
	\label{eq:feature-adapter}
\end{equation}
where $A_{\mathrm{feat}}$ maps the concatenated input to $\mathbb{R}^{C}$ and $\mathrm{vec}(\cdot)$ flattens the rotation matrix.

\subsection{Association-Decoupled Cooperative Integration}
\label{sec:decoupled-integration}

\method{} separates evidence admission from identity commitment.
After the first decoder block and temporal update,
CSA places cooperative evidence in available slots without
assigning track IDs.
URA refines the augmented bank through the remaining layers.
Only after final decoding does CPIA bind admitted hypotheses
to identities using refined ego states and persistent
correspondences, independently of CSA allocation weights.

\paragraph{Confidence-ranked Slot Admission (CSA).}
\label{sec:slot}
CSA selects cooperative evidence and determines its placement
without assigning ego-track IDs.
In-region queries $\widetilde{Q}_t^{\mathrm{in}}$ are ranked
by cooperative confidence, with budget $K$ limiting
the selected set $\mathcal{J}_t$.
Let $\mathcal{H}_t$ and $\mathcal{F}_t^0$ index live tracks
and free slots in $B_t^{\mathrm{temp}}$.
To restrict admission to free slots away from live-track
centers, eligible slots satisfy
\begin{equation}
	\mathcal{F}_t
	=
	\left\{
	i\in\mathcal{F}_t^0:
	\min_{h\in\mathcal{H}_t}
	\|p_i-p_h\|_{\mathrm{BEV}}\geq\tau_{\mathrm{CSA}}
	\right\},
	\label{eq:csa-free-slots}
\end{equation}
where $p_i$ and $p_h$ are centers at the bank-update stage.
All free slots are eligible when $\mathcal{H}_t$ is empty.
Admission is skipped if the candidate or eligible-slot set is empty.

During training, each selected query distributes its
contribution across eligible slots according to distance:
\begin{equation}
	\pi_{ji}
	=
	\frac{\exp(-d_{ji}/T_{\mathrm{CSA}})}
	{\sum_{i'\in\mathcal{F}_t}\exp(-d_{ji'}/T_{\mathrm{CSA}})},
	\qquad
	d_{ji}=\|\tilde{p}_j^{\mathrm{coop}}-p_i\|_{\mathrm{BEV}}.
	\label{eq:csa-soft}
\end{equation}
Both datasets normalize weights over slots without an
occupancy dustbin.
For slots with incoming mass $\rho_i>\epsilon_{\rho}$,
contributions are aggregated as
\begin{equation}
	\rho_i=\sum_{j\in\mathcal{J}_t}\pi_{ji},
	\qquad
	\bar{f}_i^{\mathrm{coop}}
	=\frac{\sum_{j\in\mathcal{J}_t}\pi_{ji}
		\tilde{f}_j^{\mathrm{coop}}}{\rho_i},
	\qquad
	\bar{p}_i^{\mathrm{coop}}
	=\frac{\sum_{j\in\mathcal{J}_t}\pi_{ji}
		\tilde{p}_j^{\mathrm{coop}}}{\rho_i}.
	\label{eq:csa-soft-state}
\end{equation}
Soft initialization adds the aggregated feature and
averages the slot and cooperative anchor centers:
\begin{equation}
	f_i\leftarrow f_i+\bar{f}_i^{\mathrm{coop}},
	\qquad
	a_i^{xyz}\leftarrow
	\tfrac{1}{2}a_i^{xyz}
	+\tfrac{1}{2}\bar{p}_i^{\mathrm{coop}}.
	\label{eq:csa-soft-write}
\end{equation}
Other anchor components remain unchanged, and slots with
$\rho_i\leq\epsilon_{\rho}$ retain their states.

At inference, candidates are processed in descending
confidence order.
Each overwrites the feature and complete anchor of its
nearest unused eligible slot, until either set is exhausted.
CSA retains each admitted query's source ID and assigned
slot $i(j)$ for post-decoding identity binding,
allocation itself does not determine the ego-track ID.

	\begin{figure*}[t]
		\centering
		\includegraphics[width=\textwidth]{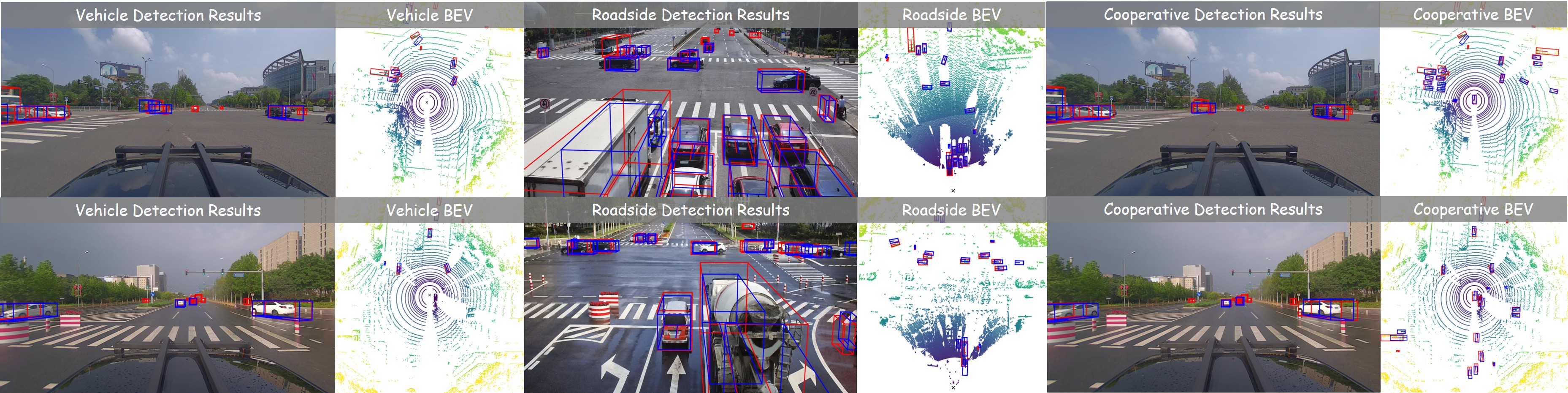}
		\caption{
			\textbf{Cooperative detection on V2X-Seq.}
			Each row shows vehicle-only, infrastructure-only,
			and cooperative results in paired image and BEV views.
			Blue: predictions, red: ground truth.
		}
		\label{fig:visualization}
		\vspace{-5pt}
	\end{figure*}
	
	\begin{figure*}[t]
		\centering
		\includegraphics[width=\linewidth]{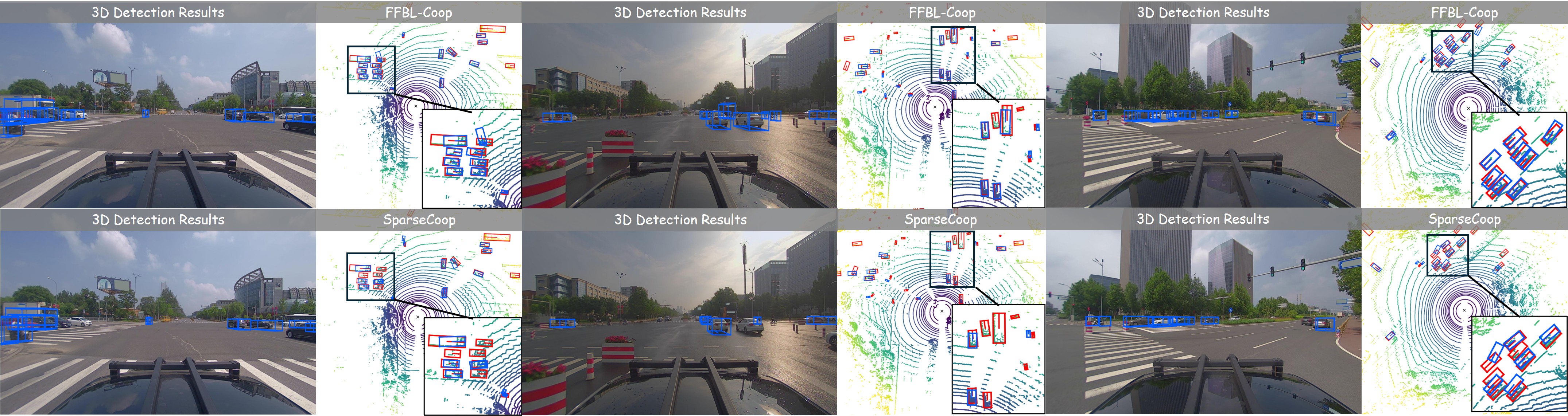}
		\caption{
			\textbf{Comparison with SparseCoop on V2X-Seq.}
			Top: \method{}, bottom: SparseCoop.
			Image and BEV pairs include enlarged regions highlighting
			detection differences.
		}
		\label{fig:qualitative}
		\vspace{-18pt}		
	\end{figure*}

\paragraph{Unified Representation Aggregation (URA).}
\label{sec:ura}

Cooperative and ego features may encode different visible parts
of the same object. URA uses cooperative semantics to guide
ego-feature retrieval: CSA-initialized queries enter the
shared decoder, with anchors locating ego-image sampling
regions and query states conditioning deformable aggregation:
\begin{equation}
	\bigl(f_i^{\ell+1},a_i^{\ell+1}\bigr)
	=
	\mathcal{D}_{\ell}
	\bigl(f_i^{\ell},a_i^{\ell},F_t,B_t^{\mathrm{temp}}\bigr).
	\label{eq:decoder-pass}
\end{equation}
Here, $F_t$ contains ego spatial features.
This query-conditioned spatial readout gathers ego evidence
beyond the information already pooled into instance vectors
by direct linear or mean fusion.
Admitted and native ego queries are refined in place through
the remaining decoder layers, producing final states
$(\hat{f}_i,\hat{a}_i)$ for CPIA identity association.

Let $\Omega_t$ denote slots initialized with cooperative instance
representations. A complementary local readout supplies
nearby cooperative context to queries outside $\Omega_t$:
\begin{equation}
	c_i=\mathrm{Attn}\!\left(
	f_i,\{\tilde{f}_j^{\mathrm{coop}}\}_{j\in\mathcal{N}_{k_{\mathrm{nbr}}}(i)}
	\right),
	\qquad
	f_i\leftarrow f_i+g_i\odot c_i,
	\quad i\notin\Omega_t.
	\label{eq:local-readout}
\end{equation}
Here, $\mathcal{U}_t$ denotes cooperative query indices eligible for local readout, and
$\mathcal{N}_{k_{\mathrm{nbr}}}(i)\subseteq\mathcal{U}_t$
contains their nearest $k_{\mathrm{nbr}}$ neighbors within radius $r$.
The gate is
$g_i=\mathrm{sigmoid}(\mathrm{MLP}([f_i;c_i;d_i^{\min}]))$,
where $d_i^{\min}$ is the nearest valid cooperative query distance.
Empty neighborhoods skip the update. In-place decoding and local readout use the same forward operations during training and inference.

\paragraph{Cooperative-Priority Identity Anchoring (CPIA).}
\label{sec:idbind}
CPIA resolves identities after final decoding, independently
of CSA allocation weights.
Association uses a fixed snapshot of final ego features
and centers, binding changes identity ownership while
preserving decoded features and boxes.

\textit{Learned association.}
An MLP predicts logits $S_{jh}$ from cooperative-query
features, final ego-track features $\hat{f}_h$,
and their spatial relations.
For a scoring set of $m$ cooperative queries and
$n_{\mathrm{live}}$ candidate ego tracks,
we append an unmatched logit:
\begin{equation}
	A=\mathrm{softmax}_{\mathrm{row}}
	\left(
	\left[
	S/T_{\mathrm{id}}
	\;\middle|\;
	b_{\mathrm{dust}}\mathbf{1}_m
	\right]
	\right),
	\label{eq:cpia-affinity}
\end{equation}
where $T_{\mathrm{id}}=0.2$ and $b_{\mathrm{dust}}=0$.
For $m>1$, we initialize
$v^{(0)}=\mathbf{1}_{n_{\mathrm{live}}+1}$
and apply $20$ scaling iterations:
\begin{equation}
	u^{(q)}=\mathbf{1}_m\oslash(Av^{(q-1)}),
	\qquad
	v^{(q)}=\mathbf{1}_{n_{\mathrm{live}}+1}
	\oslash(A^\top u^{(q)}),
	\label{eq:cpia-scaling}
\end{equation}
where $\oslash$ denotes element-wise division.
For $m=1$, we retain $A$ to avoid degenerate column scaling.
For $m>1$,
unmatched queries compete for the unmatched column's
unit total mass:
\begin{equation}
	P=
	\begin{cases}
		A, & m=1,\\
		\operatorname{diag}(u^{(20)})A
		\operatorname{diag}(v^{(20)}), & m>1,
	\end{cases}
	\qquad
	\mathcal{L}_{\mathrm{id}}
	=-\sum_{j=1}^{m}\log P_{j,y_j}.
	\label{eq:cpia-sinkhorn-ce}
\end{equation}
The target $y_j$ indexes the corresponding live track,
or the unmatched column $n_{\mathrm{live}}+1$.

\textit{Persistent identity binding.}
At inference, CPIA labels admitted hypotheses in their
assigned slots $i(j)$.
Map $M$ stores cooperative-to-ego correspondences from
new-track initialization or accepted association.
Valid mappings take priority if the mapped track is live
and its ID is unclaimed by other admitted hypotheses.
For unresolved queries and live tracks with unclaimed IDs,
CPIA computes $P$, excluding the unmatched column
from pair selection without using its score for rejection.
Pairs are greedily accepted in descending score order,
requiring unused rows and columns and
$\|\tilde{p}_j^{\mathrm{coop}}-\hat{p}_h\|_{\mathrm{BEV}}
\leq\tau_{\mathrm{id}}$,
where $\hat{p}_h$ is the final ego-track center.
Association is skipped if either set is empty
or $\tau_{\mathrm{id}}=0$.
Associated hypotheses inherit matched track IDs,
those still without an identity receive new IDs.
On ID transfer, the previous slot relinquishes the ID
and stops output and propagation under it.
Assignments update $M$, retaining one owner per ID
in the next-frame bank.

\subsection{Training Objectives}
\label{sec:loss}

Initial cooperative training starts from pretrained single-agent
checkpoints with randomly initialized cooperative modules.
On both datasets, cooperative modules
and all six decoder layers are trained, with the backbone
and FPN frozen. The overall objective is
\begin{equation}
	\mathcal{L}
	=
	\sum_{\ell=1}^{L}
	\left[
	\mathcal{L}_{\mathrm{cls}}^{(\ell)}
	+
	\mathcal{L}_{\mathrm{reg}}^{(\ell)}
	+
	\lambda_{\mathrm{dn}}
	\left(
	\mathcal{L}_{\mathrm{cls,dn}}^{(\ell)}
	+
	\mathcal{L}_{\mathrm{reg,dn}}^{(\ell)}
	\right)
	\right]
	+
	\lambda_{\mathrm{id}}\mathcal{L}_{\mathrm{id}},
	\label{eq:loss}
\end{equation}
where classification, box regression, and denoising follow the single-agent recipe. The identity loss in Eq.~\ref{eq:cpia-sinkhorn-ce} is applied once per frame with $\lambda_{\mathrm{id}}=1.0$.

\par
\endgroup
	
\section{Experiments}
\label{sec:experiments}

We evaluate \method{} in vehicle-to-infrastructure (V2I) and vehicle-to-drone (V2D) cooperation. Experiments examine overall performance, module contributions,
gains by camera-projection coverage, communication efficiency,
and robustness to calibration errors and latency.

\paragraph{Datasets and Metrics.}
We use V2X-Seq-SPD~\citep{yu2023v2xseq} under its official $2\,\mathrm{Hz}$ validation protocol~\citep{hao2025} and the Griffin-25M subset~\citep{wang2025griffin} for V2D evaluation. Both benchmarks use nuScenes-style evaluation~\citep{caesar2020nuscenes}. We report car-class AMOTA and AP, with AP averaged over center-distance thresholds $\{0.5,1,2,4\}\,\mathrm{m}$. Ablations additionally report identity switches (IDS), and communication cost is measured in bytes per second (B/s). Evaluation and latency-injection protocols are detailed in Secs.~\ref{sec:supp_protocol} and~\ref{sec:supp_latency_protocol}.

\paragraph{Implementation Details.}
\method{} uses MMDetection3D with a ResNet-50-FPN backbone and a six-layer decoder of hidden dimension $C=256$. Training uses AdamW with learning rate $3\times10^{-4}$, weight decay $0.01$, and batch size $8$. Roadside features use a $0.4\,\mathrm{m}$ BEV grid, while Drone image features use FPN strides $\{4,8,16\}$. Main results use codebook-compressed features.
Module ablations follow the initialization and training
protocol specified in Sec.~\ref{sec:supp_protocol}.
All six decoder layers are trained on both datasets, with the backbone
and FPN frozen.
Replacement rules are specified in Sec.~\ref{sec:ablation}. Baseline results in Table~\ref{tab:sota} are taken from \citet{zhong2025cooptrack,wang2025sparsecoop} under their original system configurations. No Fusion denotes the published non-cooperative baseline. Implementation details are provided in Sec.~\ref{sec:supp_protocol}.

\begin{table*}[t]
	\centering
	\small
	\belowrulesep=0pt
	\aboverulesep=0pt
	\setlength{\tabcolsep}{8pt}
	\renewcommand{\arraystretch}{1.05}
	\caption{
		System-level comparison on V2X-Seq and Griffin-25M.
		AP and AMOTA are in percent, TC is transmission cost in B/s.
		\sgdetpp{} uses $K_{\mathrm{code}}=256/512$, respectively.
		Baselines retain their published architectures and results.
		$^{\dagger}$Long-SCOPE reports the $0$--$50\,\mathrm{m}$
		subset of its long-range evaluation and is included for reference.
		Bold marks the best results excluding this reference row.
	}
	\label{tab:sota}
	\begin{tabular}{
			c|
			cc@{\hspace{10pt}}c@{\hspace{10pt}}|
			cc@{\hspace{10pt}}c@{\hspace{10pt}}
		}
		\toprule
		\multirow{2}{*}{Method}
		& \multicolumn{3}{c|}{V2X-Seq}
		& \multicolumn{3}{c}{Griffin-25M} \\
		\cmidrule(lr){2-4}
		\cmidrule(lr){5-7}
		& AP$\uparrow$ & AMOTA$\uparrow$ & TC$\downarrow$
		& AP$\uparrow$ & AMOTA$\uparrow$ & TC$\downarrow$ \\
		\midrule
		No Fusion
		& 16.6 & 13.0 & $0$
		& 37.5 & 36.5 & $0$ \\
		Late Fusion
		& 19.6 & 26.3 & $6.60\times10^{2}$
		& 37.8 & 37.7 & $1.56\times10^{3}$ \\
		Early Fusion
		& 24.3 & 20.9 & $8.19\times10^{7}$
		& 60.7 & 67.0 & $3.11\times10^{8}$ \\
		\midrule
		V2X-ViT (ECCV 22)
		& 26.8 & 28.7 & $2.56\times10^{6}$
		& 46.5 & 50.8 & $8.00\times10^{5}$ \\
		UniV2X (AAAI 25)
		& 29.5 & 23.9 & $6.96\times10^{4}$
		& 41.9 & 45.6 & $5.58\times10^{4}$ \\
		CoopTrack (ICCV 25)
		& 39.0 & 32.8 & $5.64\times10^{4}$
		& 47.9 & 48.8 & $1.17\times10^{5}$ \\
		SparseCoop (AAAI 26)
		& 53.0 & 42.1 & $3.17\times10^{4}$
		& 55.9 & 50.9 & $9.73\times10^{4}$ \\
		Long-SCOPE (CVPR 26)$^{\dagger}$
		& 53.2 & 57.6 & --
		& 59.7 & 56.6 & -- \\
		\midrule
		\textbf{\sgdetpp{} (Ours)}
		& \textbf{54.8} & \textbf{61.1} & $8.48\times10^{3}$
		& \textbf{65.3} & \textbf{68.8} & $3.37\times10^{4}$ \\
		\bottomrule
	\end{tabular}
\end{table*}

\subsection{Main Results}
\label{sec:comparison}

Table~\ref{tab:sota} compares complete systems in detection, tracking, and communication. Baseline results use their published architectures and training recipes. Sec.~\ref{sec:ablation} assesses component contributions through retrained variants sharing perception initialization and a common training protocol.

On V2X-Seq, \method{} achieves AP/AMOTA of $54.8/61.1$, exceeding SparseCoop's reported results by $1.8/19.0$ points. On Griffin-25M, it reaches $65.3/68.8$, with gains of $9.4/17.9$ points. These results demonstrate improved detection and aggregate tracking performance on both evaluated benchmarks.
With codebook compression, \method{} has estimated payload rates
of $8.48\times10^3/3.37\times10^4\,\mathrm{B/s}$
on V2X-Seq/Griffin-25M, respectively, corresponding to
approximately $26.8\%/34.6\%$ of SparseCoop's reported rates. On Griffin-25M, it exceeds Early Fusion's AMOTA of $67.0$ with substantially less communication. Sec.~\ref{sec:efficiency} examines the effect of compression.

\subsection{Ablation Study}
\label{sec:ablation}

All variants in Table~\ref{tab:ablation} follow
Sec.~\ref{sec:supp_protocol}.
Cooperative modules and all six decoder layers are trained
on both datasets, with the backbone and FPN frozen.
Without CSA, allocation uses BEV-distance-based Hungarian
matching under the same budget and gate.
Without URA, mean feature aggregation replaces in-place
refinement.
Without CPIA, slot-ID inheritance replaces learned association
and persistent mapping.
Base combines these replacements while retaining cooperative input.

\begin{table*}[ht]
	\centering
	\small
	\belowrulesep=0pt
	\aboverulesep=0pt
	\setlength{\tabcolsep}{8.8pt}
	\renewcommand{\arraystretch}{1.05}
	\caption{
		Module ablations. Disabled CSA, URA, and CPIA use Hungarian
		allocation, mean aggregation, and slot-ID inheritance, respectively.
		URA-only uses one Hungarian assignment for admission and
		identity inheritance. Base retains cooperative input.
		Bold marks the best displayed results.
	}
	\label{tab:ablation}
	\begin{tabular}{c|ccc|ccc|ccc}
		\toprule
		\multirow{2}{*}{Base}
		& \multirow{2}{*}{CSA}
		& \multirow{2}{*}{URA}
		& \multirow{2}{*}{CPIA}
		& \multicolumn{3}{c|}{V2X-Seq}
		& \multicolumn{3}{c}{Griffin-25M} \\
		\cmidrule(lr){5-7}
		\cmidrule(lr){8-10}
		& & & &
		AP$\uparrow$ & AMOTA$\uparrow$ & IDS$\downarrow$
		& AP$\uparrow$ & AMOTA$\uparrow$ & IDS$\downarrow$ \\
		\midrule
		$\checkmark$ & & &
		& 39.9 & 47.7 & 123
		& 48.7 & 48.4 & 162 \\
		\midrule
		$\checkmark$ & & $\checkmark$ &
		& 53.2 & 50.4 & 116
		& 64.0 & 55.3 & 86 \\
		$\checkmark$ & & $\checkmark$ & $\checkmark$
		& 54.2 & 60.1 & \textbf{69}
		& 64.1 & 66.1 & 57 \\
		$\checkmark$ & $\checkmark$ & & $\checkmark$
		& 53.2 & 59.0 & 72
		& 60.6 & 64.9 & \textbf{48} \\
		$\checkmark$ & $\checkmark$ & $\checkmark$ &
		& 54.3 & 59.3 & 78
		& 62.9 & 67.3 & 124 \\
		\midrule
		$\checkmark$ & $\checkmark$ & $\checkmark$ & $\checkmark$
		& \textbf{54.8} & \textbf{61.1} & 70
		& \textbf{65.3} & \textbf{68.8} & 50 \\
		\bottomrule
	\end{tabular}
\end{table*}

\paragraph{Module Contributions and Identity Association.}
Full achieves the highest AP and AMOTA among the displayed configurations on both datasets, improving over Base by $14.9/16.6$ AP and $13.4/20.4$ AMOTA points on V2X-Seq/Griffin-25M. Removing URA causes the largest leave-one-out drops: $1.6/4.7$ AP and $2.1/3.9$ AMOTA points. Removing CPIA lowers AMOTA by $1.8/1.5$ points and increases IDS from $70/50$ to $78/124$. Under Hungarian allocation, adding CPIA to URA-only raises AMOTA from $50.4/55.3$ to $60.1/66.1$ and reduces IDS from $116/86$ to $69/57$, while AP changes from $53.2/64.0$ to $54.2/64.1$. With allocation and URA architecture fixed, this supports CPIA's combined correspondence establishment and persistent reuse beyond slot-ID inheritance. Replacing Hungarian allocation with CSA yields Full, adding $1.0/2.7$ AMOTA points. CPIA improves tracking under both allocation rules, although Full does not minimize IDS. Fixed-checkpoint interventions on V2X-Seq yield AMOTA of $60.5$ for Map-only, $61.1$ for Full, and $32.5$ for Association-only (Sec.~\ref{sec:supp_inference_checks}). These results support retaining established correspondences across frames, with learned association providing an additional gain under this protocol.

Table~\ref{tab:visibility_split} compares Full and Concat-only
using the same fixed model.
Positive-depth projections define ego-only, coop-only,
and both-agent bins. 
Both-agent AP rises from $39.2/42.2$ to $53.5/72.1$
on V2X-Seq/Griffin-25M, while AP in the other bins remains nearly unchanged.
The strongest gains thus occur in shared camera coverage,
which accounts for approximately $22\%$ of evaluated objects
on V2X-Seq versus $74\%$ on Griffin-25M.
This smaller fraction, together with the smaller within-bin
gain, helps explain the lower overall detection improvement
on V2X-Seq.

\vspace{-2pt}

\subsection{Robustness and Communication Efficiency}
\label{sec:robustness}

\vspace{-2pt}

\begin{wraptable}{r}{0.6\columnwidth}
	\vspace{-10pt}
	\centering
	\small
	\belowrulesep=0pt
	\aboverulesep=0pt
	\renewcommand{\arraystretch}{1.05}
	
	\setlength{\tabcolsep}{3.3pt}
	\caption{
		AP by camera-projection coverage.
		$N$: ground-truth count. Concat-only: no in-region fusion.
	}
	\label{tab:visibility_split}
	\begin{tabular}{l|ccc|ccc}
		\toprule
		\multirow{2}{*}{Bin}
		& \multicolumn{3}{c|}{V2X-Seq}
		& \multicolumn{3}{c}{Griffin-25M} \\
		\cmidrule(lr){2-4}
		\cmidrule(lr){5-7}
		& $N$ & Concat-only & Full
		& $N$ & Concat-only & Full \\
		\midrule
		ego-only
		& 1673 & 34.9 & \textbf{35.0}
		& 178 & \textbf{18.6} & 17.7 \\
		coop-only
		& 2699 & 62.9 & \textbf{63.1}
		& 1569 & 54.9 & \textbf{55.6} \\
		both
		& 1227 & 39.2 & \textbf{53.5}
		& 4959 & 42.2 & \textbf{72.1} \\
		\bottomrule
	\end{tabular}
	
	\par\vspace{6pt}
	
	\setlength{\tabcolsep}{3.5pt}
	\caption{
		Calibration sensitivity on V2X-Seq.
	}
	\label{tab:robust_calib}
	\begin{tabular}{l|ccc|ccc}
		\toprule
		\multirow{2}{*}{Perturbation}
		& \multicolumn{3}{c|}{\sgdetpp{}}
		& \multicolumn{3}{c}{SparseCoop} \\
		\cmidrule(lr){2-4}
		\cmidrule(lr){5-7}
		& AP$\uparrow$ & AMOTA$\uparrow$ & IDS$\downarrow$
		& AP$\uparrow$ & AMOTA$\uparrow$ & IDS$\downarrow$ \\
		\midrule
		None
		& 54.8 & 61.1 & 70
		& 53.0 & 42.1 & 74 \\
		Yaw $0.5^{\circ}$
		& 53.3 & 61.0 & 72
		& 48.8 & 41.8 & 87 \\
		Yaw $1.0^{\circ}$
		& 49.4 & 60.4 & 67
		& 45.7 & 40.0 & 88 \\
		Trans. $0.5\,\mathrm{m}$
		& 50.1 & 59.9 & 68
		& 46.6 & 40.9 & 85 \\
		\bottomrule
	\end{tabular}
	
	\par\vspace{6pt}
	
	\setlength{\tabcolsep}{4pt}
	\caption{
		Communication latency on Griffin-25M.
	}
	\label{tab:robust_latency}
	\begin{tabular}{l|ccc|ccc}
		\toprule
		\multirow{2}{*}{Method}
		& \multicolumn{3}{c|}{AP$\uparrow$}
		& \multicolumn{3}{c}{AMOTA$\uparrow$} \\
		\cmidrule(lr){2-4}
		\cmidrule(lr){5-7}
		& 0\,ms & 100\,ms & 200\,ms
		& 0\,ms & 100\,ms & 200\,ms \\
		\midrule
		Early Fusion
		& 60.7 & 56.4 & 50.8
		& 67.0 & 64.0 & 58.0 \\
		SparseCoop
		& 55.9 & 55.3 & 54.0
		& 50.9 & 49.0 & 48.6 \\
		\textbf{Ours}
		& \textbf{65.3} & \textbf{61.6} & \textbf{59.4}
		& \textbf{68.8} & \textbf{68.6} & \textbf{68.4} \\
		\bottomrule
	\end{tabular}
	
	\par\vspace{6pt}
	
	\setlength{\tabcolsep}{6.8pt}
	\caption{
		Feature-compression trade-offs on V2X-Seq.
		B/s denotes estimated query payload rate.
	}
	\label{tab:codebook}
	\begin{tabular}{l|cccc}
		\toprule
		Codebook & B/s$\downarrow$
		& AP$\uparrow$ & AMOTA$\uparrow$ & IDS$\downarrow$ \\
		\midrule
		Uncompressed
		& $1.72\times10^{5}$ & 54.9 & 61.2 & 49 \\
		$K_{\mathrm{code}}=1024$
		& $8.64\times10^{3}$ & 54.7 & 59.4 & 59 \\
		$\boldsymbol{K_{\mathrm{code}}=256}$
		& $8.48\times10^{3}$ & 54.8 & 61.1 & 70 \\
		$K_{\mathrm{code}}=16$
		& $8.48\times10^{3}$ & 54.8 & 59.2 & 63 \\
		Box only
		& $8.32\times10^{3}$ & 50.6 & 50.2 & 199 \\
		\bottomrule
	\end{tabular}
	
	\vspace*{-20pt}
\end{wraptable}

\paragraph{Calibration Perturbations.}
Table~\ref{tab:robust_calib} shows that at $1.0^\circ$ yaw error, \sgdetpp{} loses $5.4$ AP points
but only $0.7$ AMOTA points, while IDS remains close
to the unperturbed level.
Translation errors show a similar pattern.
Across the tested conditions, \sgdetpp{} retains higher AP
and AMOTA and fewer IDS than SparseCoop.
These results show that degraded spatial alignment does not cause a comparable drop in tracking performance in the full pipeline.

\vspace{-2pt}

\paragraph{Communication Latency.}
Table~\ref{tab:robust_latency} evaluates Griffin-25M
with velocity compensation enabled.
At $200\,\mathrm{ms}$ delay, \sgdetpp{} loses $5.9$ AP points
but only $0.4$ AMOTA points.
Thus, stale observations reduce detection accuracy,
while aggregate tracking performance remains comparatively stable.
\sgdetpp{} retains higher AP and AMOTA than the displayed
baselines at all tested delays.
Its smaller AMOTA drop than SparseCoop, despite a larger
AP drop, further illustrates this distinction.

\vspace{-2pt}

\paragraph{Communication Efficiency.}
\label{sec:efficiency}
On the fixed Full V2X-Seq model (Table~\ref{tab:codebook}),
$K_{\mathrm{code}}=256$ reduces estimated payload from
$1.72\times10^5$ to $8.48\times10^3\,\mathrm{B/s}$,
with $0.1$-point drops in AP and AMOTA and IDS increasing
from $49$ to $70$.
Even $16$ codewords yield AP/AMOTA of $54.8/59.2$,
versus $50.6/50.2$ without feature transmission,
supporting the value of cooperative semantic features
for cooperative aggregation.
Both $16$ and $256$ codewords use one-byte indices,
giving equal steady-state payloads, while $256$
achieves higher AMOTA.
Accounting includes all transmitted queries
before receiver-side filtering.

\vspace{-6pt}

\section{Conclusion}
\label{sec:conclusion}

\vspace{-6pt}

In this work, we present \sgdetpp{}, a \emph{fuse first, bind later}
framework that separates cooperative evidence admission
from identity inheritance.
CSA allocates cooperative evidence, URA guides ego-feature retrieval
with cooperative semantics, and CPIA combines learned association
with persistent ID mappings to bind identities after refinement.
Together, these stages integrate complementary observations
while supporting tracking continuity across changes in
query placement.
On V2X-Seq and Griffin-25M, \sgdetpp{} improves detection
and tracking with compact feature transmission.
Ablations support decoupled identity management,
while perturbation studies show comparatively stable
tracking despite reduced detection accuracy.
Future work will explore uncertainty-aware association
and identity coordination across multiple agents under
changing visibility and connectivity.

\section*{AI Use Statement}
Generative AI tools were used to aid or polish writing, for retrieval and discovery (e.g., finding related work), and for research ideation or execution (e.g., reviewer-style feedback). Details are described in the paper. The authors verified all citations, technical descriptions, analyses, and revisions, and take responsibility for the final content of this work.

	\bibliography{iclr2027}
	\bibliographystyle{iclr2027}

\appendix
\newpage
\section*{Appendix}

\section{Overview}
\label{sec:appendix_overview}

This appendix provides detailed evaluation outputs,
qualitative examples, additional ablations, and
reproducibility settings for \method{}.
These analyses complement the main results by examining
module contributions, operating conditions, and
communication efficiency.

Unless stated, results are reported for the \texttt{car} class using the benchmark evaluators. Occlusion and range analyses reuse predictions with centre-distance thresholds $\{0.5,1,2,4\}\,\mathrm{m}$. Ground-truth objects outside the evaluated bin are ignored, and predictions matched to ignored objects are excluded from scoring. Metric scales and directions are specified in the corresponding tables.

The supplement is organized as follows:
\begin{itemize}
	\item Section B reports detailed evaluation
	outputs on V2X-Seq-SPD and Griffin-25M, including
	threshold-specific AP and MOTA across recall levels.
	
	\item Section C presents qualitative comparisons
	on V2X-Seq and Griffin-25M.
	
	\item Section D examines performance by
	occlusion severity and object range.
	
	\item Section~\ref{sec:supp_ablation_design} provides
	complete module ablations and additional controlled analyses.
	
	\item Section~\ref{sec:sup_robustness} reports
	robustness and communication-efficiency analyses.
	
	\item Section~\ref{sec:supp_protocol} specifies
	training, evaluation, and communication-accounting protocols.
	
	\item Section~\ref{sec:notes_limitations} discusses
	methodological distinctions and limitations.
\end{itemize}

\section{Official Evaluation Records}
\label{sec:released_records}

Tables~\ref{tab:detection_metrics} and~\ref{tab:tracking_metrics}
provide detailed evaluator outputs for the Full checkpoints
reported in Table~\ref{tab:sota}.
Both use codebook-compressed features, with
$K_{\mathrm{code}}=256$ on V2X-Seq and $512$ on Griffin-25M.
All results concern the car class.
AP, MOTA, and AMOTA are reported in percent.

\begin{table}[ht]
	\centering
	\small
	\belowrulesep=0.2pt
	\aboverulesep=0pt
	\renewcommand{\arraystretch}{1.05}
	\setlength{\tabcolsep}{12pt}
	\caption{
		Detection results at centre-distance thresholds
		of $0.5$, $1$, $2$, and $4\,\mathrm{m}$.
		Overall AP averages the four threshold-specific scores.
		ATE and AVE are measured in $\mathrm{m}$ and
		$\mathrm{m/s}$, respectively.
	}
	\label{tab:detection_metrics}
	\begin{tabular}{l|cccc|ccc}
		\toprule
		Dataset
		& AP@0.5 & AP@1 & AP@2 & AP@4
		& AP$\uparrow$ & ATE$\downarrow$ & AVE$\downarrow$ \\
		\midrule
		V2X-Seq
		& 19.95 & 49.95 & 70.05 & 79.25
		& 54.8 & 0.504 & 1.093 \\
		Griffin-25M
		& 44.15 & 66.35 & 73.05 & 77.65
		& 65.3 & 0.291 & 1.661 \\
		\bottomrule
	\end{tabular}
\end{table}

\begin{table}[ht]
	\centering
	\small
	\belowrulesep=0.2pt
	\aboverulesep=0pt
	\renewcommand{\arraystretch}{1.05}
	\setlength{\tabcolsep}{5.2pt}
	\caption{
		Tracking results and transmitted payload rates.
		TP, FP, FN, and IDS use a $2\,\mathrm{m}$ matching
		threshold. MOTA subscripts denote recall levels.
		Communication accounting follows Table~\ref{tab:sota}.
	}
	\label{tab:tracking_metrics}
	\begin{tabular}{l|c|rrrr|cccc}
		\toprule
		Dataset & B/s
		& TP & FP & FN & IDS
		& MOTA$_{0.3}$ & MOTA$_{0.5}$ & MOTA$_{0.7}$
		& AMOTA$\uparrow$ \\
		\midrule
		V2X-Seq
		& $8.48\times10^3$
		& 3510 & 551 & 2000 & 70
		& 27.7 & 45.5 & 52.9 & 61.1 \\
		Griffin-25M
		& $3.37\times10^4$
		& 4982 & 669 & 1790 & 50
		& 30.9 & 46.2 & 60.8 & 68.8 \\
		\bottomrule
	\end{tabular}
\end{table}

\paragraph{Detection and Localization.}
Table~\ref{tab:detection_metrics} reports detection accuracy
at different localization tolerances.
On V2X-Seq, AP rises from $19.95$ at $0.5\,\mathrm{m}$
to $79.25$ at $4\,\mathrm{m}$. On Griffin-25M,
it rises from $44.15$ to $77.65$.
The gains with relaxed thresholds indicate that precise
localization remains a limitation.
ATE and AVE complement overall AP by separately
characterizing center and velocity estimation errors.

\paragraph{Tracking Errors.}
Table~\ref{tab:tracking_metrics} shows that false negatives
exceed FP and IDS combined on each dataset.
Missed objects therefore remain a substantial source
of error at the reported operating points, highlighting
the importance of object recovery alongside identity
continuity.
MOTA at the selected recall levels complements AMOTA
by showing performance at individual operating points.
Table~\ref{tab:ablation} further examines the contributions
of admission, representation refinement, and identity
management.

\section{Additional Qualitative Evidence}
\label{sec:supp_qualitative}

\begin{figure}[ht]
	\centering
	\begin{subfigure}[t]{\textwidth}
		\includegraphics[width=\textwidth]{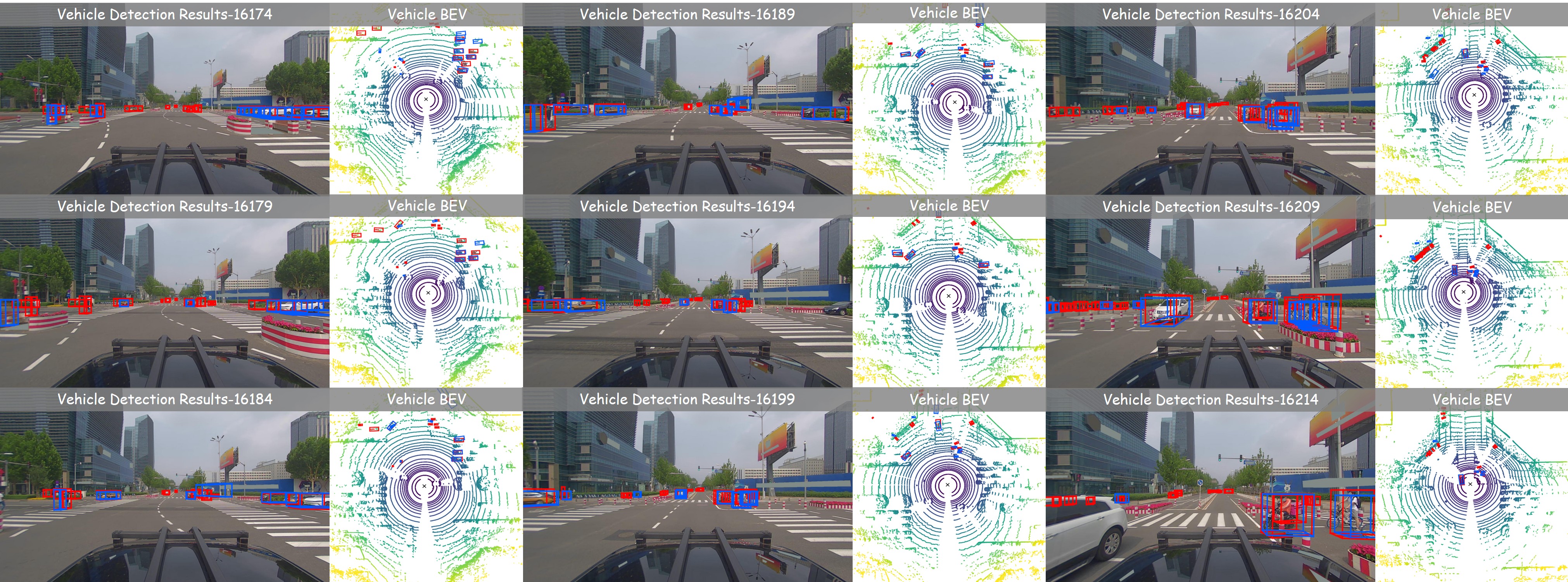}
		\caption{Ego-only nine-frame sequence (vehicle tokens $016174$--$016214$). Red: ground truth, blue: detections.}
		\label{fig:sup_veh}
	\end{subfigure}
	
	\vspace{0.4em}
	
	\begin{subfigure}[t]{\textwidth}
		\includegraphics[width=\textwidth]{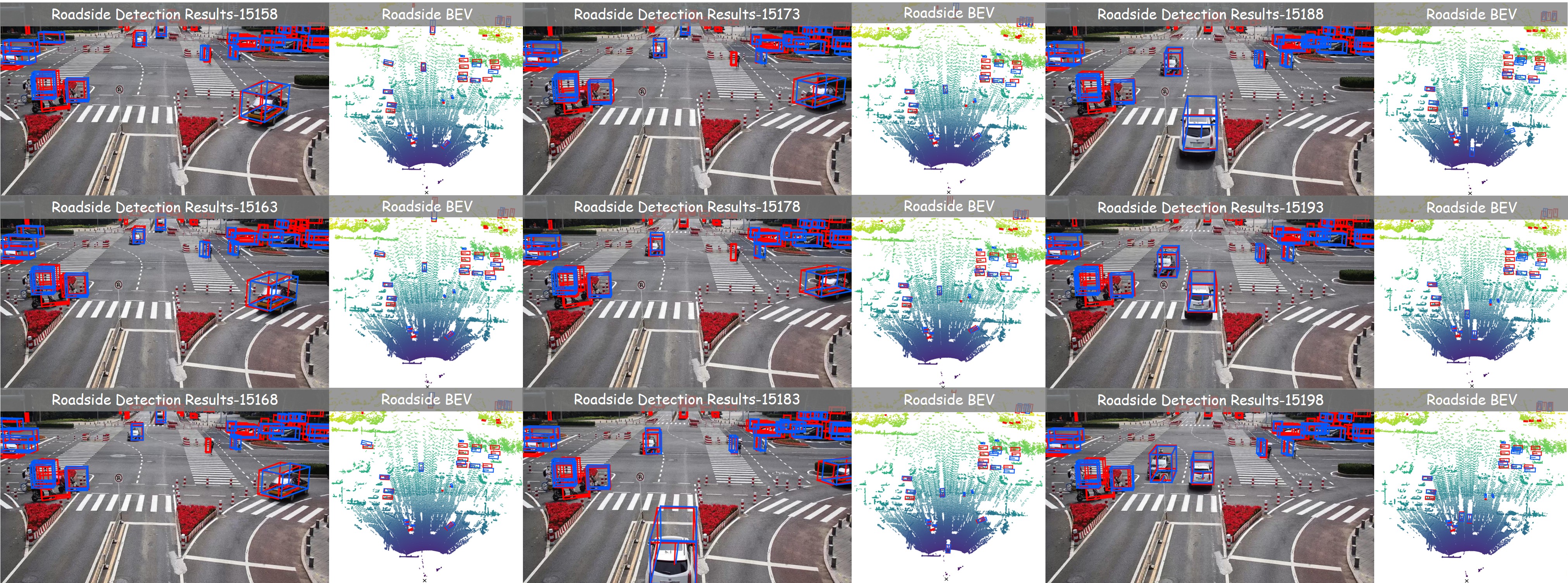}
		\caption{Roadside-only nine-frame sequence (infrastructure tokens $015158$--$015198$, aligned with vehicle tokens). }
		\label{fig:sup_inf}
	\end{subfigure}
	
	\vspace{0.4em}
	
	\begin{subfigure}[t]{\textwidth}
		\includegraphics[width=\textwidth]{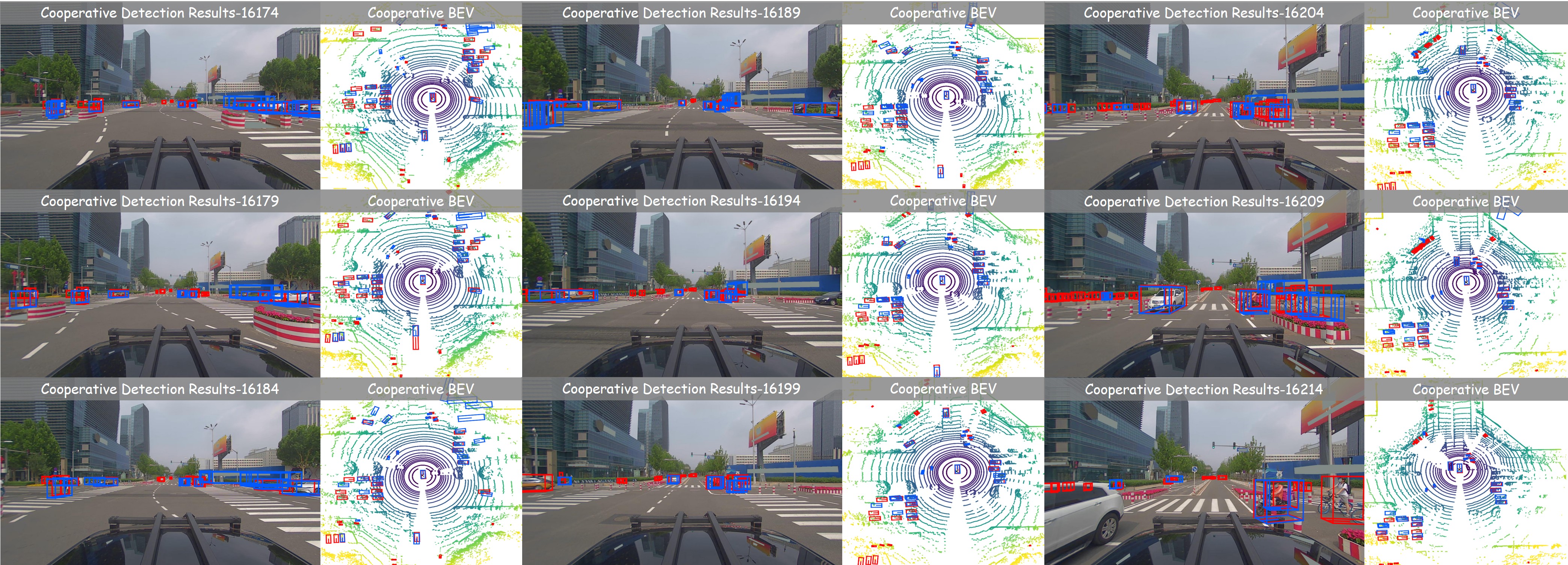}
		\caption{Cooperative nine-frame sequence on the same vehicle tokens. }
		\label{fig:sup_coop}
	\end{subfigure}
	\vspace{-5pt}
	
	\caption{Comparison of ego-only, roadside-only, and cooperative detection sequences on V2X-Seq. }
	\label{fig:sup_all}
	\vspace{-12pt}
\end{figure}

\begin{figure}[ht]
	\centering
	\begin{subfigure}[t]{\textwidth}
		\includegraphics[width=\textwidth]{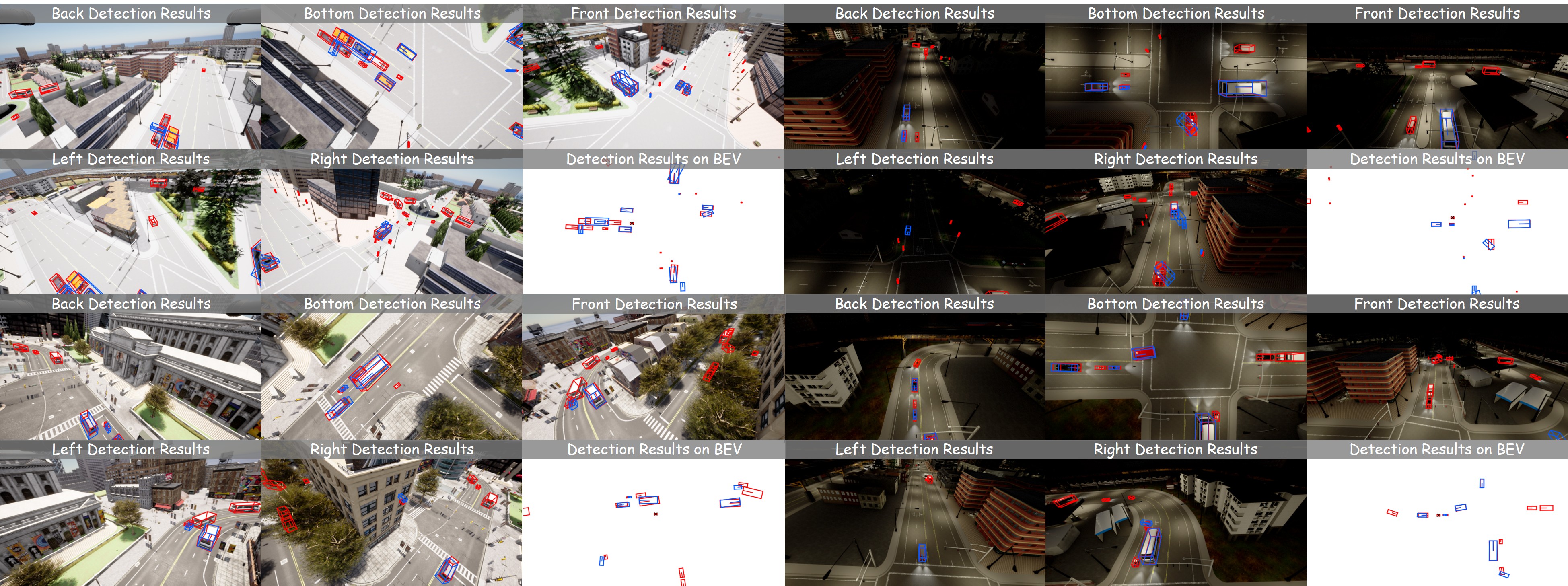}
		\caption{Drone-side detections on Griffin-25M. Red: ground truth, blue: detections.}
		\label{fig:sup_griffin_drone}
	\end{subfigure}
	
	\vspace{0.4em}
	
	\begin{subfigure}[t]{\textwidth}
		\includegraphics[width=\textwidth]{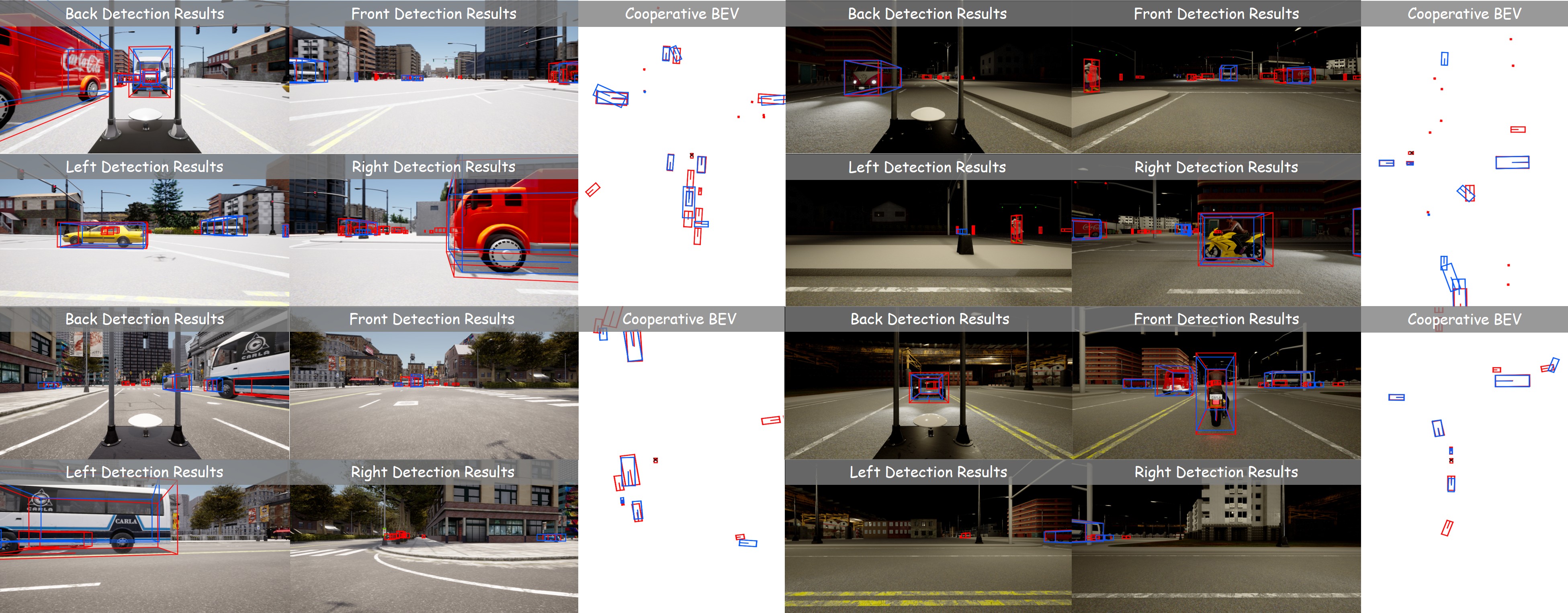}
		\caption{Vehicle-side cooperative detections on the same Griffin-25M samples. }
		\label{fig:sup_griffin_veh}
	\end{subfigure}
	
	\caption{Cooperative detections on Griffin-25M from both drone and ego-vehicle perspectives. }
	\label{fig:sup_griffin}
	\vspace*{-10pt}
\end{figure}

\paragraph{Vehicle--Infrastructure Cooperation.}
Figure~\ref{fig:sup_all} compares ego-only, roadside-only,
and cooperative predictions over nine aligned
V2X-Seq frames at $2\,\mathrm{Hz}$.
Vehicle tokens span $016174$--$016214$, with infrastructure
tokens $015158$--$015198$.
Roadside predictions include cars near the crosswalk and
far-right region that are absent from the ego-only outputs.
The cooperative sequence illustrates how these complementary
observations extend detection coverage at the intersection.

\paragraph{Vehicle--Drone Cooperation.}
Figure~\ref{fig:sup_griffin} presents four Griffin-25M
validation samples from drone and ego-vehicle perspectives.
Each sample pairs a camera view with a BEV visualization,
with ground truth in red and detections in blue.
The paired views illustrate how viewpoint differences
change object appearance and visibility.
Cooperative outputs show detections of objects that appear
small or partially occluded in ego images.

\section{Occlusion and Range Stratified Evaluation}
\label{sec:supp_visibility}

We examine detection performance across occlusion levels
and object ranges using stored predictions without retraining.
Base uses the Concat-only configuration: it disables
in-region cooperative fusion while retaining out-of-region
cooperative box concatenation. Full enables the complete
cooperative pipeline.
These comparisons measure the benefit of in-region
cooperative integration over box concatenation.
Ground-truth objects outside the active bin are ignored,
and predictions matched to ignored objects are excluded
from scoring. AP and recall are reported in percent.

\subsection{Occlusion Protocol}
\label{sec:supp_occlusion_protocol}

For a ground-truth object $i$, let
$\mathcal{V}_i^{\mathrm{ego}}$ denote the ego cameras
in which its box has a valid projection with positive
center depth. For each camera $c$, let $P_i^c$ denote
the image-clipped convex hull of the projected box
and $z_i^c$ its center depth.
For objects with nonzero projected area, we compute
the projected-occlusion ratio as
\begin{equation}
	o_i =
	\frac{
		\sum_{c\in\mathcal{V}_i^{\mathrm{ego}}}
		\left|
		P_i^c \cap
		\left(
		\bigcup_{j:z_j^c<z_i^c} P_j^c
		\right)
		\right|
	}{
		\sum_{c\in\mathcal{V}_i^{\mathrm{ego}}}
		|P_i^c|
	}.
	\label{eq:supp_occlusion_ratio}
\end{equation}
Only closer ground-truth boxes contribute to the overlap,
with all areas measured in pixels.
Objects are grouped as light ($o_i<0.2$) or occluded
($o_i\geq0.2$).
This geometric measure approximates occlusion through
projected box overlap.

\begin{table}[t]
	\centering
	\small
	\belowrulesep=0.2pt
	\aboverulesep=0pt
	\renewcommand{\arraystretch}{1.05}
	\setlength{\tabcolsep}{14pt}
	\caption{
		Car detection by occlusion severity.
		$N$ counts ground-truth objects.
		Base uses concat-only input.
		AP and recall are in percent, and $\Delta$ denotes
		the AP gain in percentage points.
		R@2m is Full recall at a $2\,\mathrm{m}$ matching threshold.
	}
	\label{tab:supp_occ_split}
	\begin{tabular}{l|c|cccc}
		\toprule
		Dataset / bin & $N$
		& Base AP & Full AP & $\Delta$ & Full R@2m \\
		\midrule
		V2X-Seq / light
		& 2981 & 64.6 & 68.8 & $+4.2$ & 95.2 \\
		V2X-Seq / occluded
		& 2618 & 33.3 & 38.4 & $+5.1$ & 84.0 \\
		\midrule
		Griffin-25M / light
		& 6528 & 35.5 & 56.4 & $+20.9$ & 70.5 \\
		Griffin-25M / occluded
		& 1683 & 0.9 & 39.2 & $+38.3$ & 75.0 \\
		\bottomrule
	\end{tabular}
\end{table}

\paragraph{Occlusion-Stratified Results.}
Table~\ref{tab:supp_occ_split} shows AP improvements
in both occlusion bins on both datasets.
On V2X-Seq, Full improves AP by $4.2$ points for light
occlusion and $5.1$ points for occluded objects.
On Griffin-25M, the corresponding gains are $20.9$
and $38.3$ points.
The larger gain in Griffin's occluded bin indicates
a substantial benefit from in-region cooperative integration
beyond the retained box-concatenation path.

\subsection{Range-Stratified Results}
\label{sec:supp_range_analysis}

Table~\ref{tab:supp_range_split} evaluates Base and Full
predictions across ego-frame distance bins, examining
where cooperative input improves detection and where
precise localization remains challenging.

\begin{table}[t]
	\centering
	\small
	\belowrulesep=0.2pt
	\aboverulesep=0pt
	\renewcommand{\arraystretch}{1.05}
	\setlength{\tabcolsep}{15pt}
	\caption{
		Car detection by ego-frame range.
		$N$ counts ground-truth objects.
		AP is in percent, and $\Delta$ denotes
		the AP gain over concat-only Base in percentage points.
		Full@0.5 reports AP at a $0.5\,\mathrm{m}$
		centre-distance threshold.
	}
	\label{tab:supp_range_split}
	\begin{tabular}{l|c|cccc}
		\toprule
		Dataset / range & $N$
		& Base AP & Full AP & $\Delta$ & Full@0.5 \\
		\midrule
		V2X-Seq / $0$--$20\,\mathrm{m}$
		& 1193 & 58.5 & 59.8 & $+1.3$ & 29.6 \\
		V2X-Seq / $20$--$40\,\mathrm{m}$
		& 2751 & 53.4 & 57.6 & $+4.2$ & 24.2 \\
		V2X-Seq / $40$--$50\,\mathrm{m}$
		& 1655 & 33.6 & 40.3 & $+6.7$ & 10.9 \\
		\midrule
		Griffin-25M / $0$--$20\,\mathrm{m}$
		& 3742 & 31.6 & 47.3 & $+15.7$ & 33.7 \\
		Griffin-25M / $20$--$40\,\mathrm{m}$
		& 3136 & 22.3 & 61.1 & $+38.8$ & 40.3 \\
		Griffin-25M / $40$--$50\,\mathrm{m}$
		& 1333 & 3.2 & 27.6 & $+24.4$ & 9.6 \\
		\bottomrule
	\end{tabular}
\end{table}

\begin{figure}[ht]
	\centering
	\includegraphics[width=\textwidth]{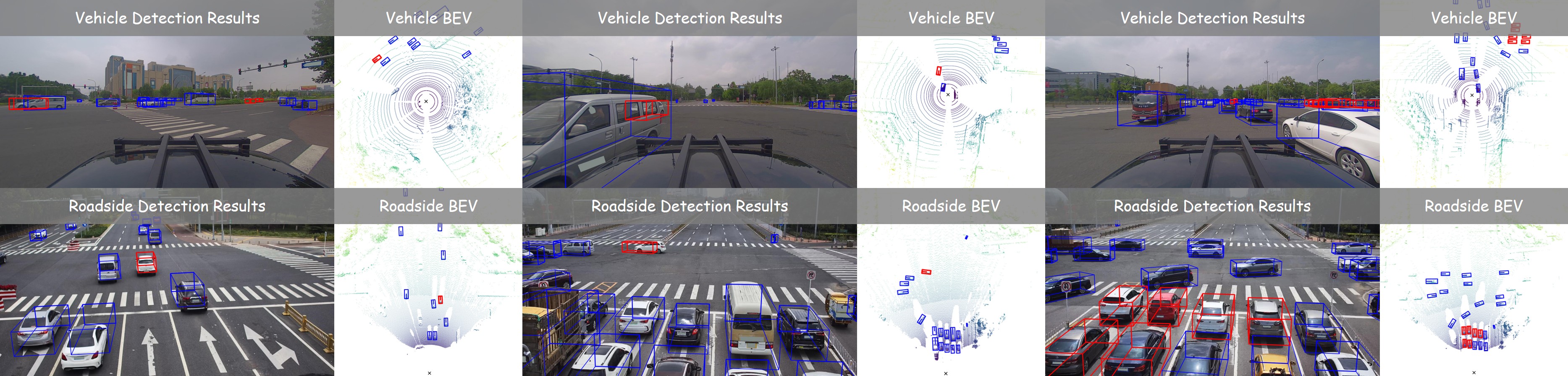}
	\caption{
		Paired ego and roadside views on V2X-Seq.
		Nearby vehicles obscure objects in the ego view,
		while the roadside perspective provides
		complementary observations.
	}
	\label{fig:sup_occ}
\end{figure}

Full improves AP over Base in every evaluated range bin,
showing that the detection benefits extend across
nearby and distant objects.
At $40$--$50\,\mathrm{m}$, AP increases from $33.6$ to
$40.3$ on V2X-Seq and from $3.2$ to $27.6$ on Griffin-25M.
Nevertheless, Full@0.5 remains at $10.9$ and $9.6$,
respectively, indicating that precise localization
remains challenging in this range.
Figure~\ref{fig:sup_occ} illustrates the complementary
coverage available from roadside observations, which
reveal objects obscured in the ego view.
This provides a qualitative example of useful cooperative
evidence, while the range-stratified results establish
detection gains within each bin without isolating
their causes.

\section{Additional Ablation Studies}
\label{sec:supp_ablation_design}

We extend the main ablations with complete module combinations
and additional analyses of admission, refinement, and identity
management. Module variants follow the initialization and training
protocol in Sec.~\ref{sec:supp_protocol}.
Fixed-checkpoint diagnostics are identified separately.

\subsection{Complete Module Ablations}

Table~\ref{tab:ablation2} reports eight combinations sharing
the codebook and training schedule. Cooperative modules and all six decoder layers are trained on both datasets, with the backbone and FPN frozen. Disabled CSA, URA, and CPIA use Hungarian
allocation, mean aggregation without local readout, and
slot-ID inheritance, respectively. Base retains cooperative
input with these replacements.

\begin{table*}[!ht]
	\centering
	\small
	\belowrulesep=0pt
	\aboverulesep=0pt
	\setlength{\tabcolsep}{8.8pt}
	\renewcommand{\arraystretch}{1.0}
	\caption{
		Complete module ablations under a shared training
		protocol. Base uses the stated replacements.
		Bold marks the best results.AP and AMOTA are reported in percent.
	}
	\label{tab:ablation2}
	\begin{tabular}{c|ccc|ccc|ccc}
		\toprule
		\multirow{2}{*}{Base}
		& \multirow{2}{*}{CSA}
		& \multirow{2}{*}{URA}
		& \multirow{2}{*}{CPIA}
		& \multicolumn{3}{c|}{V2X-Seq}
		& \multicolumn{3}{c}{Griffin-25M} \\
		\cmidrule(lr){5-7}
		\cmidrule(lr){8-10}
		& & & &
		AP$\uparrow$ & AMOTA$\uparrow$ & IDS$\downarrow$
		& AP$\uparrow$ & AMOTA$\uparrow$ & IDS$\downarrow$ \\
		\midrule
		$\checkmark$ & & &
		& 39.9 & 47.7 & 123
		& 48.7 & 48.4 & 162 \\
		\midrule
		$\checkmark$ & $\checkmark$ & &
		& 52.6 & 56.5 & 73
		& 59.2 & 63.4 & \textbf{43} \\
		$\checkmark$ & & $\checkmark$ &
		& 53.2 & 50.4 & 116
		& 64.0 & 55.3 & 86 \\
		$\checkmark$ & & & $\checkmark$
		& 53.0 & 58.9 & 75
		& 60.5 & 63.8 & 66 \\
		\midrule
		$\checkmark$ & & $\checkmark$ & $\checkmark$
		& 54.2 & 60.1 & \textbf{69}
		& 64.1 & 66.1 & 57 \\
		$\checkmark$ & $\checkmark$ & & $\checkmark$
		& 53.2 & 59.0 & 72
		& 60.6 & 64.9 & 48 \\
		$\checkmark$ & $\checkmark$ & $\checkmark$ &
		& 54.3 & 59.3 & 78
		& 62.9 & 67.3 & 124 \\
		\midrule
		$\checkmark$ & $\checkmark$ & $\checkmark$ & $\checkmark$
		& \textbf{54.8} & \textbf{61.1} & 70
		& \textbf{65.3} & \textbf{68.8} & 50 \\
		\bottomrule
	\end{tabular}
\end{table*}

\paragraph{Individual Contributions.}
The single-module controls reveal different effects
on detection and tracking.
URA yields the highest AP among these variants,
whereas CPIA yields the highest AMOTA on each dataset.
CSA alone also improves both metrics over Base.
These patterns suggest that representation quality
and identity handling contribute to aggregate
tracking performance.

\paragraph{Combined Effects.}
The complete combinations show that module gains
depend on the surrounding allocation and identity configuration.
On Griffin-25M, adding URA to CSA raises AMOTA from
$63.4$ to $67.3$ but increases IDS from $43$ to $124$,
with CPIA also enabled, AMOTA reaches $68.8$ and IDS
decreases to $50$.
Thus, improved aggregate tracking accuracy can coexist
with more identity switches, motivating separate evaluation
of overall tracking performance and identity continuity.
The combinations also show non-monotonic detection
effects: adding CSA to URA-only lowers AP from
$64.0$ to $62.9$, whereas adding it to URA+CPIA raises
AP from $64.1$ to $65.3$.
The modules provide complementary benefits
without uniformly additive improvements across configurations.

\subsection{Hyperparameter Sensitivity}
\label{sec:supp_hyperparams}

Table~\ref{tab:ablation_hyper} varies admission budget $K$
and CPIA association radius $\tau_{\mathrm{id}}$
individually on fixed Full checkpoints for V2X-Seq and Griffin-25M,
with all other inference settings unchanged.
The default budgets are $35$ and $50$, respectively.
The default CPIA radii and CSA thresholds are
$2.2\,\mathrm{m}$ on V2X-Seq and $1.5\,\mathrm{m}$ on Griffin-25M.
The CSA thresholds remain fixed throughout these sweeps
to distinguish association-radius sensitivity from allocation gating.

\begin{table*}[!ht]
	\centering
	\small
	\belowrulesep=0pt
	\aboverulesep=0pt
	\renewcommand{\arraystretch}{1.0}
	\setlength{\tabcolsep}{13.5pt}
	\caption{
		Inference-time sensitivity on V2X-Seq.
		All removes the budget cap.
		$\tau_{\mathrm{id}}=0$ disables spatial adoption.
		AP and AMOTA are in percent. Bold marks the defaults.
	}
	\label{tab:ablation_hyper}
	\begin{tabular}{l|ccc|l|ccc}
		\toprule
		\multicolumn{4}{c|}{Admission budget}
		& \multicolumn{4}{c}{CPIA adoption radius} \\
		\cmidrule(lr){1-4}
		\cmidrule(lr){5-8}
		$K$ & AP$\uparrow$ & AMOTA$\uparrow$ & IDS$\downarrow$
		& $\tau_{\mathrm{id}}$
		& AP$\uparrow$ & AMOTA$\uparrow$ & IDS$\downarrow$ \\
		\midrule
		5 & 53.6 & 58.7 & 37
		& $0\,\mathrm{m}$ & 54.7 & 60.5 & 76 \\
		20 & 54.6 & 61.2 & 51
		& $1.0\,\mathrm{m}$ & 54.8 & 61.1 & 72 \\
		\textbf{35}
		& \textbf{54.8} & \textbf{61.1} & \textbf{70}
		& $\mathbf{2.2\,\mathrm{m}}$
		& \textbf{54.8} & \textbf{61.1} & \textbf{70} \\
		All & 54.7 & 61.0 & 79
		& $3.0\,\mathrm{m}$ & 54.6 & 59.3 & 59 \\
		\bottomrule
	\end{tabular}
\end{table*}

\begin{table*}[!ht]
	\centering
	\small
	\belowrulesep=0pt
	\aboverulesep=0pt
	\renewcommand{\arraystretch}{1.0}
	\setlength{\tabcolsep}{13.5pt}
	\caption{
		Inference-time sensitivity on Griffin-25M.
		All removes the budget cap.
		$\tau_{\mathrm{id}}=0$ disables spatial adoption.
		AP and AMOTA are in percent. Bold marks the defaults.
	}
	\label{tab:ablation_hyper_griffin}
	\begin{tabular}{l|ccc|l|ccc}
		\toprule
		\multicolumn{4}{c|}{Admission budget}
		& \multicolumn{4}{c}{CPIA adoption radius} \\
		\cmidrule(lr){1-4}
		\cmidrule(lr){5-8}
		$K$ & AP$\uparrow$ & AMOTA$\uparrow$ & IDS$\downarrow$
		& $\tau_{\mathrm{id}}$
		& AP$\uparrow$ & AMOTA$\uparrow$ & IDS$\downarrow$ \\
		\midrule
		10 & 64.3 & 67.9 & 55
		& $0\,\mathrm{m}$ & 65.2 & 68.7 & 63 \\
		25 & 64.9 & 68.3 & 53
		& $0.75\,\mathrm{m}$ & 65.3 & 68.8 & 59 \\
		\textbf{50} & \textbf{65.3} & \textbf{68.8} & \textbf{50}
		& $\mathbf{1.5\,\mathrm{m}}$
		& \textbf{65.3} & \textbf{68.8} & \textbf{50} \\
		All & 65.3 & 68.8 & 51
		& $3.0\,\mathrm{m}$ & 65.3 & 67.9 & 53 \\
		\bottomrule
	\end{tabular}
\end{table*}

\paragraph{Admission Budget.}
On V2X-Seq, increasing $K$ from $5$ to $20$ improves
AP/AMOTA from $53.6/58.7$ to $54.6/61.2$.
Further expansion offers little aggregate benefit:
the default $K=35$ gives slightly higher AP,
while $K=20$ achieves slightly higher AMOTA and fewer IDS.
On Griffin-25M, AP and AMOTA improve as $K$ increases
from $10$ to the default $50$, with IDS decreasing
from $55$ to $50$.
Removing the cap provides no further AP or AMOTA gain
over the default on either dataset.
These results support a bounded admission budget:
admitting more candidates does not necessarily improve
detection or tracking.

\paragraph{Identity Association Radius.}
AP and AMOTA vary little across the tested radii
up to the default on each dataset.
Disabling learned association retains map reuse and
new-track initialization, yielding AMOTA of $60.5/68.7$
on V2X-Seq/Griffin-25M, compared with $61.1/68.8$
at the defaults.
Enabling association also reduces IDS from $76/63$
to $70/50$.
Increasing the radius to $3.0\,\mathrm{m}$ lowers AMOTA
to $59.3/67.9$, providing no aggregate tracking benefit.
IDS nevertheless decreases on V2X-Seq at this radius,
so its variation alone does not determine the preferred
setting.
These sweeps show modest additional gains from learned
association alongside persistent mapping, while a wider
spatial gate does not consistently improve the metrics.

\begin{table*}[t]
	\centering
	\small
	\belowrulesep=0.2pt
	\aboverulesep=0pt
	\renewcommand{\arraystretch}{1.05}
	\setlength{\tabcolsep}{16pt}
	\caption{
		Detailed ablations on V2X-Seq covering admission,
		refinement, and CPIA components.
	}
	\label{tab:supp_focused_hyperparams}
	\begin{tabular}{l|l|ccc}
		\toprule
		Component & Intervention
		& AP$\uparrow$ & AMOTA$\uparrow$ & IDS$\downarrow$ \\
		\midrule
		Reference & Full
		& 54.8 & 61.1 & 70 \\
		\midrule
		CSA & Early prefix admission
		& 50.5 & 50.9 & 67 \\
		CSA & $\tau_{\mathrm{CSA}}=3.0\,\mathrm{m}$
		& 54.5 & 60.0 & 74 \\
		\midrule
		URA & w/o local readout
		& 52.3 & 54.7 & 97 \\
		URA & Mean-vector feature replacement
		& 51.2 & 52.9 & 163 \\
		\midrule
		CPIA & Sticky occupancy
		& 54.3 & 55.3 & 103 \\
		CPIA & Map only
		& 54.7 & 60.5 & 76 \\
		CPIA & Association only
		& 54.6 & 32.5 & 167 \\
		\bottomrule
	\end{tabular}
\end{table*}

\subsection{Detailed Design Ablations}
\label{sec:supp_inference_checks}

Table~\ref{tab:supp_focused_hyperparams} examines admission,
feature integration, and identity management on V2X-Seq.
CSA and CPIA interventions use same checkpoint.
URA variants are retrained under the shared protocol.

\paragraph{Admission Timing and Spatial Gating.}
Moving admission before the temporal update reduces
AP/AMOTA by $4.3/10.2$ points, whereas increasing
$\tau_{\mathrm{CSA}}$ from $2.2$ to $3.0\,\mathrm{m}$
reduces them by $0.3/1.1$ points.
Under these interventions, the fixed model is more
sensitive to admission timing than to the tested
spatial-threshold change.

\paragraph{Cooperative Feature Integration.}
Removing local readout reduces AP/AMOTA from $54.8/61.1$
to $52.3/54.7$ and increases IDS from $70$ to $97$,
supporting its contribution beyond in-place decoding.
Replacing instance features with a mean vector yields
AP/AMOTA of $51.2/52.9$ and $163$ IDS.
Together, these retrained controls support the use of
instance-specific cooperative semantics and local context
for cooperative refinement.

\begin{figure}[ht]
	\centering
	\includegraphics[width=\textwidth]{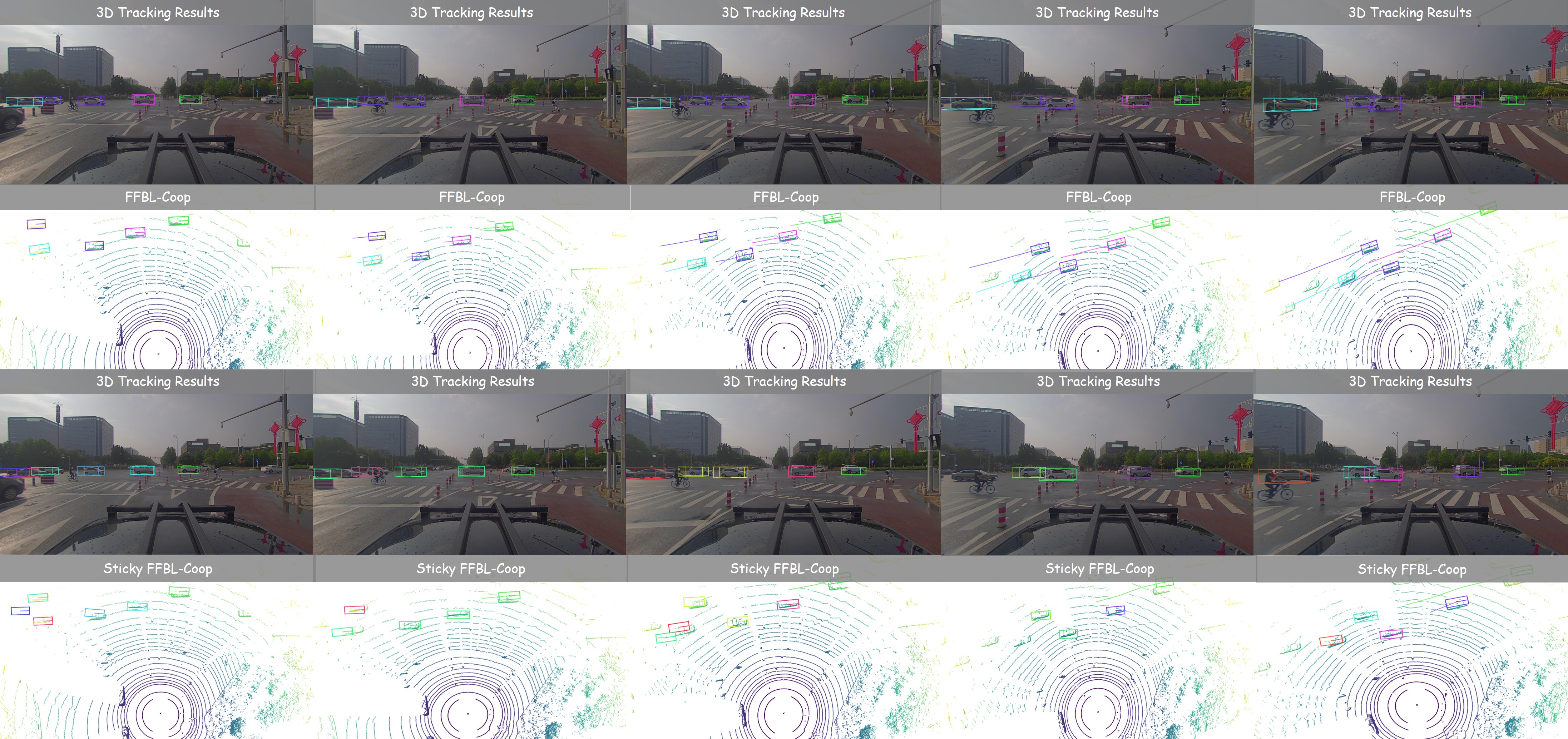}
	\caption{
		Identity continuity on five consecutive V2X-Seq frames
		(tokens $010562$--$010582$, $2\,\mathrm{Hz}$).
		Top: Full (CPIA). Bottom: CSA+URA with sticky IDs.
		Different colors denote different track IDs.
	}
	\label{fig:sup_id}
	\vspace*{-10pt}
\end{figure}

\paragraph{Correspondence Establishment and Temporal Reuse.}
Map-only yields AP/AMOTA of $54.7/60.5$ and $76$ IDS,
close to Full's $54.8/61.1$ and $70$ IDS.
Association-only retains AP at $54.6$ but lowers AMOTA
to $32.5$ and increases IDS to $167$.
Sticky occupancy yields $54.3/55.3$ and $103$ IDS.
Map-only initializes mappings through ego-ID assignment
during slot occupancy and uses them.
Full additionally associates unmapped queries with
existing ego tracks and records accepted correspondences,
adding $0.6$ AMOTA points under this protocol.
These results support persistent use of established
ID assignments, with learned association providing
an additional benefit.
Figure~\ref{fig:sup_id} illustrates the identity behavior
of Full and sticky-ID across consecutive frames.

\section{Robustness and Communication Efficiency}
\label{sec:sup_robustness}

\subsection{Latency Injection Protocol}
\label{sec:supp_latency_protocol}

On Griffin-25M, communication delay is simulated by selecting
drone observations $d$ frames earlier from the native
$10\,\mathrm{Hz}$ stream, with $d\in\{0,1,2,4\}$ corresponding
to $\{0,100,200,400\}\,\mathrm{ms}$.
Table~\ref{tab:robust_latency} reports the $0$, $100$, and
$200\,\mathrm{ms}$ settings.
Delayed observations are supplied to the $2\,\mathrm{Hz}$
ego evaluator.

Before coordinate alignment, each cooperative center is compensated
in the drone frame using
\begin{equation}
	p_j^{+}
	=
	p_j^{\mathrm{coop}}
	+
	v_j^{\mathrm{coop}}\Delta t,
	\qquad
	\Delta t=t^{\mathrm{ego}}-t^{\mathrm{coop}}.
\end{equation}
The compensated object state is then transformed into
the ego frame before cooperative integration.
All delay settings use the same fixed checkpoint with
velocity compensation enabled.

On V2X-Seq, we retain the provided timestamps and evaluate
calibration perturbations as a separate robustness condition.
Yaw and translation are perturbed separately by fixed
offsets at the magnitudes reported in
Table~\ref{tab:robust_calib}, applied consistently across
frames to $T_{\mathrm{coop}\rightarrow\mathrm{ego}}$.
Model weights and cooperative processing rules remain fixed.
SparseCoop is evaluated locally under the same
perturbation protocol using fixed model weights.

\subsection{Codebook Compression}
\label{sec:supp_griffin_codebook}

We evaluate feature compression on a fixed Griffin-25M
Full checkpoint. Codebooks are constructed offline by
applying $k$-means to training-set instance features.
Each $256$-dimensional drone feature is encoded by its
nearest codeword index and reconstructed by table lookup
at the receiver, following Eq.~\ref{eq:vq-enc}.
The accompanying object states and metadata are unchanged.

\paragraph{Initialization and Synchronization.}
The codebook is constructed offline from training-set
instance features and held fixed throughout evaluation.
Both agents use identical codebook entries; no online
codebook updates are performed.
If the codebook is transferred once before query streaming,
its float32 payload is $4K_{\mathrm{code}}C$ bytes,
amounting to $524{,}288$ bytes for
$K_{\mathrm{code}}=512$ and $C=256$.
The reported $C_{\mathrm{tx}}$ measures steady-state query
payload and excludes this one-time transfer.
Including the initial codebook payload, the average rate
over a session of $D$ seconds is
$C_{\mathrm{tx}}+4K_{\mathrm{code}}C/D$,
excluding packet headers and synchronization metadata.

\paragraph{Per-Query Payload.}
Under the accounting convention in Sec.~\ref{sec:supp_protocol},
an uncompressed query uses $1076$ bytes, comprising a
$1024$-byte feature and a $52$-byte non-feature record:
a $44$-byte state, a $4$-byte confidence score,
and a $4$-byte source track ID.
Codeword indices use one byte for $K_{\mathrm{code}}\leq256$
and two bytes for the larger evaluated codebooks,
giving $53$ and $54$ bytes per query, respectively.
The one-byte format is retained for $K_{\mathrm{code}}=1$.
All transmitted queries are counted before receiver-side
region filtering and admission.

\begin{table}[ht]
	\centering
	\small
	\belowrulesep=0.2pt
	\aboverulesep=0pt
	\renewcommand{\arraystretch}{1.05}
	\setlength{\tabcolsep}{22
		pt}
	\caption{
		Feature compression on a fixed Griffin-25M
		Full checkpoint.
		B/s measures steady-state query transmission,
		excluding one-time codebook synchronization.
		AP and AMOTA are in percent.
		Bold marks the default configuration.
	}
	\label{tab:supp_griffin_codebook}
	\begin{tabular}{l|cccc}
		\toprule
		Codebook & B/s$\downarrow$
		& AP$\uparrow$ & AMOTA$\uparrow$ & IDS$\downarrow$ \\
		\midrule
		Uncompressed
		& $6.71\times10^{5}$ & 65.5 & 68.9 & 50 \\
		$K_{\mathrm{code}}=1024$
		& $3.37\times10^{4}$ & 64.7 & 68.3 & 53 \\
		$\boldsymbol{K_{\mathrm{code}}=512}$
		& $\boldsymbol{3.37\times10^{4}}$
		& \textbf{65.3} & \textbf{68.8} & \textbf{50} \\
		$K_{\mathrm{code}}=256$
		& $3.31\times10^{4}$ & 64.3 & 66.2 & 68 \\
		$K_{\mathrm{code}}=64$
		& $3.31\times10^{4}$ & 64.1 & 66.2 & 64 \\
		$K_{\mathrm{code}}=1$
		& $3.31\times10^{4}$ & 63.3 & 65.7 & 63 \\
		\bottomrule
	\end{tabular}
\end{table}

\paragraph{Accuracy and Communication Trade-off.}
Table~\ref{tab:supp_griffin_codebook} shows that the default
codebook reduces estimated steady-state payload by approximately
$95.0\%$,
with AP/AMOTA decreasing by $0.2/0.1$ points
and IDS remaining at $50$.
Reducing $K_{\mathrm{code}}$ to $256$ lowers the payload
further but reduces AP/AMOTA to $64.3/66.2$.
With a single codeword, all cooperative features reconstruct
to the same vector, yielding AP/AMOTA of $63.3/65.7$.
These results support compact instance-specific semantic
cues for ego-feature aggregation, with the default codebook
retaining accuracy close to uncompressed transmission.

\section{Reproducibility and Evaluation Protocol}
\label{sec:supp_protocol}

\paragraph{Evaluation Protocols.}
We use official validation splits and benchmark evaluators.
V2X-Seq-SPD follows the $2\,\mathrm{Hz}$ protocol
of~\citet{hao2025,yu2023v2xseq}.
Griffin-25M~\citep{wang2025griffin} uses timestamp-aware
cooperative inputs.
Unless stated otherwise, evaluation concerns the
\texttt{car} class within $50\,\mathrm{m}$ using
nuScenes-style metrics~\citep{caesar2020nuscenes}.
AP averages centre-distance thresholds
$\{0.5,1,2,4\}\,\mathrm{m}$.
AP, AMOTA, and MOTA are in percent, IDS in counts,
and communication rates in B/s.
Table~\ref{tab:sota} reports published baseline
results~\citep{zhong2025cooptrack,wang2025sparsecoop}.
No Fusion denotes the published non-cooperative baseline.
Long-SCOPE~\citep{wang2026longscope} is a reference using
its $0$--$50\,\mathrm{m}$ evaluation subset.
Stratified analyses reuse stored predictions, excluding
out-of-bin ground truth and predictions matched to it.
Projection bins require positive camera depth.
Full and Concat-only share model weights and retain
out-of-region box concatenation. The camera-projection,
occlusion-, and range-stratified analyses compare these
configurations. In the occlusion and range tables,
Concat-only is denoted as Base.

\paragraph{Architecture and Training.}
We use MMDetection3D with a ResNet-50 backbone, four-level
FPN, and six-layer decoder with feature dimension $C=256$.
Images are resized to $544\times960$ on V2X-Seq
and $448\times800$ on Griffin-25M.
Roadside BEV resolution is $0.4\,\mathrm{m}$,
aerial image readout uses FPN strides $\{4,8,16\}$.
Training uses camera-only inputs and the classes
$\{\texttt{car},\texttt{bicycle},\texttt{pedestrian}\}$.
The backbone and FPN remain frozen, cooperative modules
and all six decoder layers are trained.
We use AdamW with learning rate $3\times10^{-4}$,
weight decay $0.01$, batch size $8$, cosine annealing,
and mixed precision on four GPUs.
Single-agent pretraining lasts $800/100$ epochs on
V2X-Seq/Griffin-25M, followed by initial cooperative training
and $4/8$ epochs of soft-admission fine-tuning.
Griffin-25M uses a $24$-epoch initial cooperative
identity-binding stage.
Final fine-tuning continues each preceding cooperative model.

\paragraph{Query Banks and Cooperative Updates.}
V2X-Seq uses $N=360$ slots, $120$ ego-reserved slots,
and admission budget $K=35$,
Griffin-25M uses $N=900$, $300$ ego-reserved slots,
and $K=50$.
CSA follows the first decoder layer and temporal update,
with $\tau_{\mathrm{CSA}}=2.2/1.5\,\mathrm{m}$
and $T_{\mathrm{CSA}}=0.5/0.3$, respectively.
Both Full configurations enable local readout with
$k_{\mathrm{nbr}}=8$, radius $r=8\,\mathrm{m}$,
and Gaussian factor
$\exp(-d^2/(2\sigma_{\mathrm{URA}}^2))$,
where $\sigma_{\mathrm{URA}}=4\,\mathrm{m}$.
CSA follows Eqs.~\ref{eq:csa-soft-state}--\ref{eq:csa-soft-write}
with $\epsilon_{\rho}=10^{-6}$.
This dimensionless threshold gates initialization,
normalization by $\rho_i$ does not attenuate accepted updates.
CPIA is supervised between cooperative queries and final
decoded states of temporally retained live ego tracks.
Ground-truth object identities determine the matched-track
or unmatched target $y_j$.
Softly aggregated slots receive detection supervision,
their mixture weights do not assign identities.
Hard slot indices, identity binding, and map updates
are used only at inference.

\paragraph{Output Routing and Identity Management.}
\label{sec:supp_integration_region}
Aligned queries are filtered in the ego frame using
$\sqrt{x^2+y^2}\leq60\,\mathrm{m}$ and $z>-3\,\mathrm{m}$,
excluding the ego footprint
$x\in[-2.04,2.04]\,\mathrm{m}$,
$y\in[-0.92,0.92]\,\mathrm{m}$.
Valid queries with positive-depth center projections
inside at least one ego image enter
$\widetilde{Q}_t^{\mathrm{in}}$ for CSA/URA.
Other valid queries follow the separate box-output path,
invalid queries are discarded.
Projection coverage does not imply absence of occlusion,
and the routing region differs from the evaluation range.
Both paths reuse valid mappings in $M$, preserving
the mapped ego ID across path changes.
The merged output contains at most one hypothesis per ego ID.
Mapping reuse requires confidence at least $0.1$,
a live and currently unclaimed ego ID, and
$t-t_{\mathrm{last}}\leq T_{\mathrm{expire}}/\alpha$.
The aging factor $\alpha\in[1,2]$ depends on confidence
and update frequency.
Base expiration times are $30/16\,\mathrm{s}$
on V2X-Seq/Griffin-25M, with reference frequencies
of $2/10\,\mathrm{Hz}$.
Expired entries fail lookup and are periodically removed.
Before each sequence, the evaluator resets the temporal
bank, identity map, and ID counter, independently
of batch-size changes.

\paragraph{Ablation Protocols.}
Within each dataset, module variants share single-agent
initialization, the codebook, and the training schedule.
Cooperative modules are randomly initialized, and each
variant completes all cooperative stages, fine-tuning
from its own preceding model.
Each variant is trained three times, selecting checkpoints
by validation AMOTA within each run.
All reported metrics come from the run with median AMOTA.
Hyperparameter, compression, and CSA/CPIA interventions
use the Full weights reported in Table~\ref{tab:sota},
changing only the specified inference conditions.
The local-readout and mean-vector controls in
Table~\ref{tab:supp_focused_hyperparams} are retrained.

\paragraph{Communication and Codebook Synchronization.}
Transmitted query payloads are counted before receiver-side
filtering or admission:
\begin{equation}
	C_{\mathrm{tx}}
	=f_{\mathrm{tx}}\bar{N}_{\mathrm{tx}}b_q
	\qquad [\mathrm{B/s}],
	\label{eq:communication-rate}
\end{equation}
where $f_{\mathrm{tx}}$ is transmission frequency,
$\bar{N}_{\mathrm{tx}}$ is mean transmitted query count,
and $b_q$ is bytes per query.
Each query contains $44$ bytes of float32 state,
$4$ bytes of confidence, and a $4$-byte source ID.
Uncompressed features require $1024$ bytes,
codeword indices use one byte for
$K_{\mathrm{code}}\leq256$ and two bytes for the tested
larger codebooks.
Total payloads are therefore $1076$, $53$, or $54$ bytes
per query, box-only transmission uses $52$ bytes.
Packet headers are excluded, and published baseline rates
retain their original accounting conventions.
V2X-Seq transmits approximately $80$ queries per frame
at $2\,\mathrm{Hz}$, yielding $8480\,\mathrm{B/s}$
with $K_{\mathrm{code}}=256$.
Griffin-25M transmits approximately $62.4$ queries per
drone frame at $10\,\mathrm{Hz}$, yielding
$3.37\times10^4\,\mathrm{B/s}$ with $K_{\mathrm{code}}=512$.
These counts differ from bank capacity and admission budget.
Codebooks are constructed offline from training-set features
and remain fixed throughout evaluation.
Reported $C_{\mathrm{tx}}$ includes query payload only.
If an initial codebook transfer is required, including
its float32 payload gives an average rate of
$C_{\mathrm{tx}}+4K_{\mathrm{code}}C/D$
over a session of $D$ seconds, excluding packet headers
and synchronization metadata.
No subsequent codebook-update traffic is incurred
in the evaluated setting.

\section{Method Notes and Limitations}
\label{sec:notes_limitations}

\paragraph{Relationship to Cooperative Query Trackers.}
CoopTrack~\citep{zhong2025cooptrack},
SparseCoop~\citep{wang2025sparsecoop}, and
Long-SCOPE~\citep{wang2026longscope} develop cross-agent
association and feature refinement using graph affinities,
geometric and appearance cues, or contextual matching.
\sgdetpp{} builds on these mechanisms through
\emph{fuse first, bind later}: CSA allocates remote evidence,
URA refines hypotheses with ego observations, and CPIA
resolves identities using persistent mappings, learned
association, and new-track initialization.
Its distinction lies in separating query placement from
identity inheritance within the recurrent pipeline,
rather than introducing unmatched-query retention
or ego-image refinement as new capabilities.

\paragraph{Dependence on Cooperative Evidence.}
Performance depends on cooperative coverage, detection
quality, and calibration.
In-region refinement relies on useful ego observations,
while out-of-region concatenation retains cooperative
localization errors.
Persistent mappings depend on consistent source IDs
and may propagate accepted identity errors.
Evaluation primarily concerns cars, broader category
coverage remains to be validated.

\paragraph{Robustness and Communication.}
Calibration and latency remain limitations despite
comparatively stable AMOTA under the tested conditions.
A $1^\circ$ yaw perturbation reduces V2X-Seq AP/AMOTA
from $54.8/61.1$ to $49.4/60.4$.
At $200\,\mathrm{ms}$ latency, Griffin-25M AP decreases
from $65.3$ to $59.4$, while AMOTA changes from
$68.8$ to $68.4$ with velocity compensation.
These evaluations assess the complete pipeline and
do not isolate individual robustness mechanisms.
Reported query payload rates exclude packet headers
and codebook synchronization, so deployment traffic
depends additionally on update frequency and protocol overhead.
Less reliable calibration, variable delays, packet loss,
and identity coordination across multiple agents
remain directions for future work.

\end{document}